\documentclass{article}

   \usepackage[final,nonatbib]{neurips_2024}

\usepackage[utf8]{inputenc} %
\usepackage[T1]{fontenc}    %
\usepackage{hyperref}       %
\usepackage{url}            %
\usepackage{booktabs}       %
\usepackage{amsfonts}       %
\usepackage{nicefrac}       %
\usepackage{microtype}      %
\usepackage{xcolor}         %
\usepackage{enumitem}

\usepackage{booktabs, makecell, tabularx}
\usepackage{multirow}
\usepackage{duckuments}
\usepackage{amsmath}
\usepackage{graphicx}
\usepackage{arydshln} %
\definecolor{MyDarkBlue}{rgb}{0,0.08,1}
\definecolor{MyDarkGreen}{rgb}{0.02,0.6,0.02}
\definecolor{MyDarkRed}{rgb}{0.8,0.02,0.02}
\definecolor{MyDarkOrange}{rgb}{0.40,0.2,0.02}
\definecolor{MyPurple}{RGB}{111,0,255}
\definecolor{MyRed}{rgb}{1.0,0.0,0.0}
\definecolor{MyGold}{rgb}{0.75,0.6,0.12}
\definecolor{MyDarkgray}{rgb}{0.66, 0.66, 0.66}
\definecolor{MyDarkCyan}{rgb}{0.05, 0.55, 0.45}
\definecolor{MyBlack}{rgb}{0., 0., 0.}
\definecolor{MyMagenta}{rgb}{1., 0., 1.}
\definecolor{BerkeleyYellow}{RGB}{255,204,41}
\definecolor{BerkeleyLightBlue}{RGB}{94,146,221}
\definecolor{BkDarkBlue}{rgb}{.05,.07,.353}

\newcommand{\arxiv}[1]{{#1}}
\newcommand{\camready}[1]{{#1}}

\newcommand{\myparagraph}[1]{\vspace{0pt} \noindent \textbf{#1} \ }

\newcommand{\etal}{\textit{et al.}}

\newcommand{\ignorethis}[1]{}

\newcommand*{\menlo}{\fontfamily{lmtt}\fontsize{9}{9}\selectfont }

\title{Data Attribution for Text-to-Image Models \\ by Unlearning Synthesized Images}

\author{Sheng-Yu Wang$^{1}$\hspace{2mm} Aaron Hertzmann$^{2}$\hspace{2mm} Alexei A. Efros$^{3}$\hspace{2mm} Jun-Yan Zhu$^{1}$\hspace{2mm} Richard Zhang$^{2}$ \\
$^{1}$Carnegie Mellon University \hspace{5mm} $^{2}$Adobe Research \hspace{5mm} $^{3}$UC Berkeley}

\begin{document}

\maketitle

\begin{abstract}
The goal of data attribution for text-to-image models is to identify the training images that most influence the generation of a new image. 
Influence is defined such that, for a given output,  if a model is retrained from scratch without the most influential images, the model would fail to reproduce the same output. 
Unfortunately, directly searching for these influential images is computationally infeasible, since it would require repeatedly retraining models from scratch. In our work,  
we propose an efficient data attribution method by simulating \textit{unlearning the synthesized image}. We achieve this by increasing the training loss on the output image, without catastrophic forgetting of other, unrelated concepts. 
We then identify training images with significant loss deviations after the unlearning process and label these as influential. We evaluate our method with a computationally intensive but ``gold-standard'' retraining from scratch and demonstrate our method's advantages over previous methods.

\end{abstract}

\section{Introduction}
\label{sec:intro}

Data attribution for text-to-image generation aims to identify which training images ``influenced'' a given output.  The black-box nature of modern image generation models~\cite{nichol2021glide,rombach2021highresolution,ramesh2022hierarchical,yu2022scaling,saharia2022photorealistic,gafni2022make,kang2023scaling,pchang2023muse}, together with the enormous datasets required \cite{schuhmann2022laion}, makes it extremely challenging to understand the contributions of individual training images. Although generative models can, at times, replicate training data~\cite{carlini2023extracting,somepalli2022diffusion}, they typically create %
samples distinct from any specific training image.  

We believe that a counterfactual definition of ``influence'' best matches the intuitive goal of attribution~\cite{georgiev2023journey,koh2017understanding}. Specifically, a collection of training images is influential for a given output image if removing those images from the training set and then retraining from scratch makes the model unable to reproduce the same synthesized image. Unfortunately, directly searching for the most influential images according to this definition is computationally infeasible since it would require training an exponentially large number of new models from scratch. 

Hence, practical influence estimation requires effective approximations. For example, many approaches replace retraining with a closed-form approximation, computed separately for each training image~\cite{georgiev2023journey,zheng2023intriguing,hampel1974influence,koh2017understanding,park2023trak}. For text-to-image attribution, these methods are outperformed by simple matching of off-the-shelf image features \cite{caron2021emerging}. Wang \etal~\cite{wang2023evaluating} use model customization~\cite{kumari2022customdiffusion} to study the effect of training a model towards an exemplar, but find limited generalization to the general large-scale training case. We aim for a tractable method that accurately predicts influence according to the counterfactual definition.

\begin{figure}
    \centering
    \includegraphics[width=\linewidth]{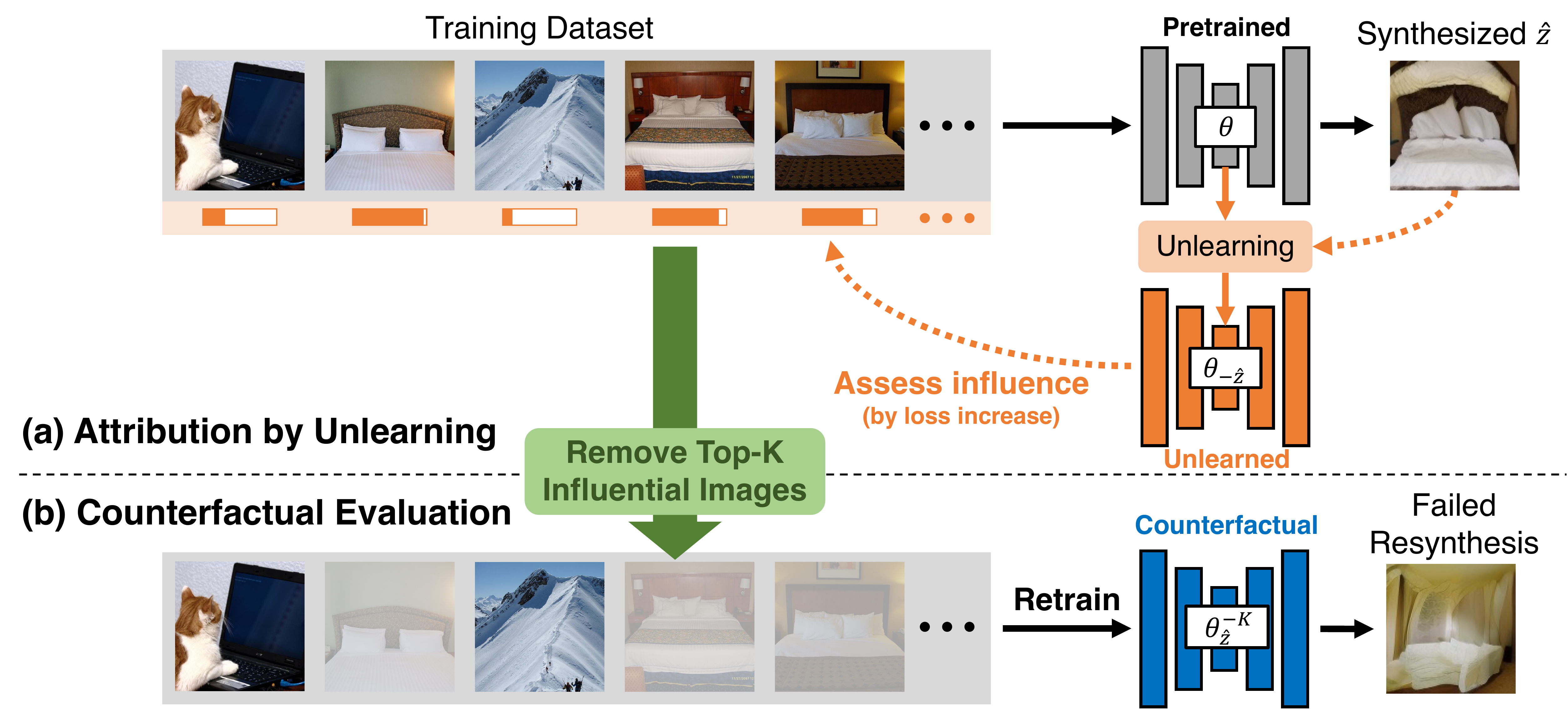}
        \vspace{-10pt}
    \caption{%
    (a) \textbf{Our algorithm}:  We propose a new data attribution method using machine unlearning.  By modifying the pretrained model $\theta$ to unlearn the synthesized result $\hat{\mathbf{z}}$, the model also forgets the influential training images crucial for generating that specific result. (b) \textbf{Evaluation}: We validate our method through counterfactual evaluation, where we retrain the model without the top $K$ influential images identified by our method. When these influential images are removed from the dataset, the model fails to generate the synthesized image.}
    \vspace{-10pt}
    \label{fig:teaser}
\end{figure}

We propose an influence prediction approach with two key ideas (Figure \ref{fig:teaser}). First, we approximate the removal of a training image from a model through an optimization procedure termed \textit{unlearning~\cite{guo2019certified,bourtoule2021machine,golatkar2020eternal},} which increases the training loss of the target image while preserving unrelated concepts. We then compute the training loss for the original synthesized image.  However, directly applying this idea would require unlearning separately for \emph{every} single training image, which is also costly. Our second main idea is to reverse the roles: we \textit{unlearn the synthesized image, and then evaluate which training images are represented worse by the new model.}  This method requires only one unlearning optimization, rather than a separate unlearning step for each training image.

The methodology for unlearning is important. Unlearning a synthesized image by naively maximizing its loss leads to catastrophic forgetting~\cite{kirkpatrick2017overcoming}, where the model also fails to generate other unrelated concepts. Inspired by work on unlearning data for classifiers~\cite{guo2019certified,tanno2022repairing}, we mitigate this issue by regularizing gradient directions using the Fisher information to retain pretrained information. Additionally, we find that updating only the key and value mappings in the cross-attention layers improves attribution performance. We show how ``influence functions'' \cite{koh2017understanding,hampel1974influence} can be understood as approximations to unlearning \arxiv{in Appendix~\ref{sec:supp_unlearning_math}}, but they are also limited by their closed-form approximation nature. %

We perform a rigorous counterfactual validation: removing a predicted set of influential images from the training set, retraining from scratch, and then checking that the synthesized image is no longer represented. We use MSCOCO~\cite{lin2014microsoft} ($\sim$100k images), which allows for retraining models within a reasonable compute budget.  We also test on a publicly-available attribution benchmark~\cite{wang2023evaluating} using customized text-to-image models~\cite{kumari2022customdiffusion}. Our experiments show that our algorithm outperforms prior work on both benchmarks, demonstrating that unlearning synthesized images is an effective way to attribute training images. Our code is available at: \url{https://peterwang512.github.io/AttributeByUnlearning}.

In summary, our contributions are:
\begin{itemize}[leftmargin=12pt]
  \setlength{\parskip}{0pt}
\item We propose a new data attribution method for text-to-image models, by unlearning the synthesized image and identifying which training images are forgotten.
\item We find and ablate the components for making unlearning efficient and effective, employing Fisher information and tuning a critical set of weights.
\item We rigorously show that our method is counterfactual predictive by omitting influential images, retraining, and checking that the synthesized image cannot be regenerated. Along with the existing Customized Model benchmark, our method identifies influential images more effectively than recent methods based on customization and influence functions.
\end{itemize}

\section{Related Work}
\label{sec:related_works}

\myparagraph{Attribution.}
\textit{Influence functions}~\cite{hampel1974influence,koh2017understanding} approximate how the objective function of a test datapoint would change after perturbing a training datapoint. One may then predict attribution according to the training points that can produce the largest changes. Koh and Liang~\cite{koh2017understanding} proposed using influence functions to understand model behavior in deep discriminative models. The influence function requires calculating a Hessian of the model parameters, for which various efficient algorithms have been proposed, such as inverse hessian-vector products~\cite{koh2017understanding}, Arnoldi iteration~\cite{schioppa2022scaling}, Kronecker factorization~\cite{martens2015optimizing,george2018fast,grosse2023studying}, \arxiv{Gauss-Newton approximation}~\cite{park2023trak}, and nearest neighbor search~\cite{guo2020fastif}.

Other methods explore different approaches. Inspired by the game-theory concept of Shapley value~\cite{shapley1953value}, several methods train models on subsets of training data and estimate the influence of a training point by comparing the models with and without that training point  %
~\cite{ghorbani2019data,jia2019towards,feldman2020neural,ilyas2022datamodels}. Pruthi et al.~\cite{pruthi2020estimating}  estimate influence by tracking train-test image gradient similarity throughout model training.

Recent methods have started tackling attribution for diffusion-based image generation. Wang~\etal~\cite{wang2023evaluating} proposed attribution by model customization~\cite{ruiz2022dreambooth,kumari2022customdiffusion}, where a pretrained model is influenced by tuning towards an exemplar concept. Several works adapt TRAK~\cite{park2023trak}, an influence function-based method, to diffusion models, extending it by attributing at specific denoising timesteps~\cite{georgiev2023journey}, or by improving gradient estimation and using Tikhonov regularization~\cite{zheng2023intriguing}. Unlike these methods, our method performs attribution by directly unlearning a synthesized image and tracking the effect on each training image. Our method outperforms existing methods in attributing both customized models and text-to-image models. \camready{Concurrent works also apply machine unlearning for attribution tasks~\cite{ko2024mirrored,isonuma2024unlearning}. Unlike their approaches, we find that applying strong regularization during unlearning is crucial to obtaining good performance in our tasks.}

\myparagraph{Machine unlearning.}
Machine unlearning seeks to efficiently ``remove'' specific training data points from a model. Recent studies have explored concept erasure for text-to-image diffusion models, specified by a text request~\cite{kumari2023conceptablation,gandikota2023erasing,gandikota2024unified,huang2023receler,zhang2023forget}, whereas we remove individual images.  While forgetting may be achieved using multiple models trained with subsets of the dataset beforehand~\cite{bourtoule2021machine,graves2021amnesiac,tarun2023fast}, doing so is prohibitively expensive for large-scale generative models.

Instead, our approach follows unlearning methods that update model weights directly~\cite{guo2019certified,nguyen2020variational,golatkar2020eternal,golatkar2021mixed,tanno2022repairing}. The majority of prior methods use the Fisher information matrix (FIM) to approximate retraining without forgetting other training points~\cite{guo2019certified,golatkar2020eternal,sekhari2021remember,tanno2022repairing,app13169341,foster2024fast}. In particular, we are inspired by the works from Guo~\etal~\cite{guo2019certified} and Tanno~\etal~\cite{tanno2022repairing}, which draw a connection between FIM-based machine unlearning methods and influence functions. We show that unlearning can be efficiently applied to the attribution problem, by ``unlearning''  output images instead of training data.

\myparagraph{Replication detection.}  
Shen \etal~\cite{shen2019discovering} identify repeated pictorial elements in art history. Somepalli \etal~\cite{somepalli2022diffusion} and Carlini \etal~\cite{carlini2023extracting} investigate the text-to-image synthesis of perceptually-exact copies of training images. Unlike these, our work focuses on data attribution for more general synthesis settings beyond replication. %

\section{Problem Setting and Evaluation}
Our goal is to attribute a generated image to its training data. 
We represent the training data as $\mathcal{D} = \{({\bf x}_i, {\bf c}_i)\}_{i=1}^{N}$, where ${\bf x} \in \mathcal{X}$ denotes an image and ${\bf c}$ represents its conditioning text. A learning algorithm $\mathcal{A}: \mathcal{D} \rightarrow \theta$ yields parameters of a generative model; for instance, $\theta = \mathcal{A}(\mathcal{D})$ is a model trained on $\mathcal{D}$. We focus on diffusion models %
that generate an image from a noise map $\epsilon \sim \mathcal{N}(0, {\bf I})$. A generated image from text $\textbf{c}$ is represented as $\hat{\textbf{x}} = G_\theta(\epsilon, \textbf{c})$. To simplify notation, we write a text-image tuple as a single entity. A synthesized pair is denoted as $\hat{\mathbf{z}} = (\hat{\textbf{x}}, \textbf{c})$, and a training pair is denoted as $\mathbf{z}_i = (\textbf{x}_i, \textbf{c}_i) \sim \mathcal{D}$.
We denote the loss of an image $\textbf{x}$ conditioned on $\textbf{c}$ as $\mathcal{L}(\mathbf{z}, \theta)$.

Next, we describe the ``gold-standard'' evaluation method that we use to define and evaluate influence. %
Section~\ref{sec:methods} describes our method for predicting influential images.
\label{sec:method_eval}
\myparagraph{Counterfactual evaluation.}
A reliable data attribution algorithm should accurately reflect a \textit{counterfactual prediction}. That is, if an algorithm can identify a set of truly influential training images, then a model trained \textit{without} those images would be incapable of generating or representing that image. As noted by Ilyas~\etal~\cite{ilyas2022datamodels} and Park~\etal~\cite{park2023trak}, counterfactual prediction is computationally intensive to validate. As such, these works introduce the Linear data modeling (LDS) score as an efficient proxy, but with the assumption that data attribution methods are \textit{additive}, which does not hold for feature matching methods and our method.

In our work, we invest substantial computational resources to the ``gold standard'' counterfactual evaluation within our resource limits. 
That is, we use an attribution algorithm to identify a critical set of $K$ images, denoted as $\mathcal{D}_{\hat{\mathbf{z}}}^K \subset \mathcal{D}$. We then train a generative model without those images from scratch, \textit{per synthesized sample and per attribution method}. Despite the computational cost,  this allows us to provide the community with a direct evaluation of counterfactual prediction,  without relying on a layer of approximations. We formalize our evaluation scheme as follows.

\myparagraph{Training a counterfactual model.} For evaluation, an attribution algorithm is given a budget of $K$ images for attributing a synthesized sample $\hat{\mathbf{z}}$, denoted as $\mathcal{D}_{\hat{\mathbf{z}}}^K$. We then train a leave-$K$-out model $\theta_{\hat{\mathbf{z}}}^{-K}$ from scratch using $\mathcal{D}_{\hat{\mathbf{z}}}^{-K} = \mathcal{D} \backslash \mathcal{D}_{\hat{\mathbf{z}}}^{K}$, the dataset with the $K$ attributed images removed:

\begin{equation}
\begin{aligned}
    \theta_{\hat{\mathbf{z}}}^{-K} &= \mathcal{A}(\mathcal{D}_{\hat{\mathbf{z}}}^{-K}),
\end{aligned}
\end{equation}

\myparagraph{Evaluating the model.} We then compare this ``leave-$K$-out'' model against $\theta_0 = \mathcal{A}(\mathcal{D})$, the model trained with the entire dataset, and assess how much it loses its capability to represent $\hat{\mathbf{z}}$ in terms of both the loss change $\Delta \mathcal{L}(\hat{\mathbf{z}}, \theta)$ and the capability to generate the same sample $\Delta G_\theta(\epsilon, \textbf{c})$ \camready{from the same input noise $\epsilon$ and text $\textbf{c}$}.

First, if the leave-$K$-out model is trained without the top influential images, it should reconstruct synthetic image $\hat{\mathbf{z}}$ more poorly, resulting in a higher $\Delta \mathcal{L}(\hat{\mathbf{z}}, \theta)$:

\begin{equation}
\begin{aligned}
    \Delta \mathcal{L}(\hat{\mathbf{z}}, \theta) = \mathcal{L}(\hat{\mathbf{z}}, \theta_{\hat{\mathbf{z}}}^{-K}) - \mathcal{L}(\hat{\mathbf{z}}, \theta_0). \\
\end{aligned}
\end{equation}

Second, if properly selected, the leave-$K$-out model should no longer be able to generate $\hat{\mathbf{x}}=G_\theta(\epsilon, \mathbf{c})$. For diffusion models, we can particularly rely on the ``seed consistency'' property~\cite{georgiev2023journey,su2022dual,khrulkov2023understanding}. %
\camready{Georgiev~\etal~\cite{georgiev2023journey} find that images generated from the same random noise have little variations, even when generated by two independently trained diffusion models on the same dataset.} They leverage this property to evaluate attribution via $\Delta G_\theta(\epsilon, \textbf{c})$, the difference of generated images between $\theta_0$ and $\theta_{\hat{\mathbf{z}}}^{-K}$. An effective attribution algorithm should lead to a leave-$K$-out model generating images that deviate more from the original images, resulting in a larger $\Delta G_\theta(\epsilon, \textbf{c})$ value:

\begin{equation}
    \Delta G_\theta(\epsilon, \textbf{c}) = d\big(G_{\theta_0}(\epsilon, \textbf{c}), G_{\theta_{\hat{\mathbf{z}}}^{-K}}(\epsilon, \textbf{c})\big),
\end{equation}

\noindent where $d$ can be any distance function, such as L2 or CLIP~\cite{radford2021learning}. \arxiv{Georgiev~\etal~\cite{georgiev2023journey} also adopt $\Delta G_\theta(\epsilon, \textbf{c})$ for evaluation.} While evaluating loss increases and seed consistency is specific to diffusion models, the overarching idea of retraining and evaluating if a synthesized image is still in the model applies across generative models.

\myparagraph{Choice of loss $\mathcal{L}(\mathbf{z}, \theta)$.}
\camready{We focus on DDPM loss introduced by Ho~\etal~\cite{ho2020denoising}, the standard loss used to train diffusion models.
Diffusion models learn to predict the noise added to a noisy image $\mathbf{x}_t = \sqrt{\bar\alpha_t}\mathbf{x}_0 + \sqrt{1 - \bar\alpha_t}\epsilon$, where $\mathbf{x}_0$ is the original image, $\epsilon$ is the Gaussian noise, $\bar\alpha_t$ and $t$ controls the noise strength. $t$ is an integer timestep sampled between 1 to $T$, where $T$ is typically set to 1000. A larger $t$ implies more noise added. DDPM loss optimizes for the noise prediction task:
$\mathbb{E}[\| \epsilon_{\theta}(\mathbf{x}_t, \mathbf{c}, t) - \epsilon\|^2],$
where $\epsilon_\theta(\cdot)$ is the denoiser for noise prediction, and $c$ denotes text condition.} %

\section{Attribution by Unlearning}
\label{sec:methods}
In this section, we introduce our attribution approach for a text-to-image model $\theta_0 = \mathcal{A}(\mathcal{D})$, trained on dataset $\mathcal{D}$.
We aim to find the highly influential images on a given synthetic image $\hat{\mathbf{z}}$ in dataset $\mathcal{D}$. %

\camready{If we had infinite computes and a fixed set of $K$ images, we could search for every possible subset of $K$ images and train models from scratch. The subset whose removal leads to the most '`forgetting'' of the synthesized image would be considered the most influential.
Of course, such a combinatoric search is impractical, so we simplify the problem by estimating the influence score of each training point \textit{individually} and selecting the top $K$ influential images based on their scores.}

\camready{Formally, we define a data attribution algorithm $\boldsymbol\tau$, which given access to the training data, model, and learning algorithm, estimates the influence of each training point, denoted by $\boldsymbol\tau(\hat{\mathbf{z}}, \mathcal{D}, \theta, \mathcal{A}) = \left[\tau(\hat{\mathbf{z}}, \mathbf{z}_1), \tau(\hat{\mathbf{z}}, \mathbf{z}_2), \dots, \tau(\hat{\mathbf{z}}, \mathbf{z}_{N}) \right]$. $\tau$ represents the factorized attribution function that estimates influence based on the synthesized sample and a training point. Although $\tau$ can access all training data, parameters, and the learning algorithm, we omit them for notational simplicity.}

\camready{One potential algorithm for $\tau(\hat{\mathbf{z}}, \mathbf{z})$ is to remove $\mathbf{z}$ from the training set, retrain a model and check its performance on $\hat{\mathbf{z}}$. However, this method is infeasible due to the size of the training set. To overcome this, we have introduced two key ideas. First, rather than training from scratch, we use model unlearning---efficiently tuning a pretrained model to remove a data point. Second, rather than unlearning training points, we apply unlearning to the \textit{synthesized} image and assess how effectively each training image is forgotten. To measure the degree of removal, we track the training loss changes for each training image after unlearning, and we find this effective for data attribution.}

\myparagraph{Unlearning the synthesized image.} 
A naive approach to unlearn a synthesized image $\hat{\mathbf{z}}$ is to solely maximize its loss $\mathcal{L}(\hat{\mathbf{z}}, \theta)$. However, only optimizing for this leads to catastrophic forgetting~\cite{kirkpatrick2017overcoming}, where the model can no longer represent other concepts.

Instead, we propose to retain the information from the original dataset while ``removing'' the synthesized image, as though it had been part of training. Given model trained on the original dataset $\theta_0 = \mathcal{A}(\mathcal{D})$ and a synthesized image $\hat{\mathbf{z}}$, we compute a new model $\theta_{-\hat{\mathbf{z}}} = \mathcal{A}(\mathcal{D} \backslash \hat{\mathbf{z}})$, with $\hat{\mathbf{z}}$ removed. Here, we use the set removal notation $\backslash$ to specify a ``negative'' datapoint in the dataset. Concretely, we solve for the following objective function, using elastic weight consolidation (EWC) loss~\cite{kirkpatrick2017overcoming} as an approximation:

\begin{equation}
\begin{aligned}
    \mathcal{L}_\text{unlearn}^{\hat{\mathbf{z}}}(\theta) = &-\mathcal{L}(\hat{\mathbf{z}}, \theta) + \sum_{\mathbf{z}\in \mathcal{D}} \mathcal{L}(\mathbf{z}, \theta) \\ 
    &\approx -\mathcal{L}(\hat{\mathbf{z}}, \theta) +  \frac{N}{2}(\theta - \theta_0)^TF(\theta - \theta_0),
\end{aligned}
\label{eqn:unlearning}
\end{equation} 
where $F$ is the Fisher information matrix, which is approximated as a diagonal form for computational efficiency. $F$ approximates the training data loss to the second-order~\cite{pascanu2013revisiting}, a technique widely used in continual learning~\cite{kirkpatrick2017overcoming,li2020fewshot}. This enables us to solve for the new model parameters  $\theta_{-\hat{\mathbf{z}}}$ efficiently, by initializing from the pretrained model $\theta_0$.
We optimize this loss with Newton updates:

\begin{equation}
\begin{aligned}
    \theta \leftarrow \theta + \frac{\alpha}{N} F^{-1}  \nabla\mathcal{L}(\hat{\mathbf{z}}, \theta),
\end{aligned}
\label{eqn:newton_removal}
\end{equation}
where $\alpha$ controls the step size, $F^{-1}$ is the inverse of the Fisher information matrix. In practice, Newton updates allow us to achieve effective attribution with few iterations and, in some cases, as few as one step. We denote the unlearned model as $\theta_{-\hat{\mathbf{z}}}$. \arxiv{We provide details of the EWC loss, and Newton update in Appendix~\ref{sec:supp_unlearning_math}.}

\myparagraph{Attribution using the unlearned model.}
After we obtain the unlearned model $\theta_{-\hat{\mathbf{z}}}$, we define our attribution function $\tau$ by tracking the training loss changes for each training sample $\mathbf{z}$:

\begin{equation}
\begin{aligned}
    \tau(\hat{\mathbf{z}}, \mathbf{z}) = \mathcal{L}(\mathbf{z}, \theta_{-\hat{\mathbf{z}}}) - \mathcal{L}(\mathbf{z}, \theta_0).
\end{aligned}
\label{eqn:attribution_score}
\end{equation}

The value $\tau(\hat{\mathbf{z}}, \mathbf{z})$ is expected to be close to zero for most unrelated training images since the EWC loss used in obtaining $\theta_{-\hat{\mathbf{z}}}$ acts as a regularization to preserve the original training dataset.
A higher value of $\tau(\hat{\mathbf{z}}, \mathbf{z})$ indicates that unlearned model $\theta_{-\hat{\mathbf{z}}}$ no longer represents the training sample $\mathbf{z}$.

\myparagraph{Relation with influence functions.} %
Our method draws a parallel with the influence function, which aims to estimate the loss change of $\hat{\mathbf{z}}$ by removing a training point $\mathbf{z}$. However, training leave-one-out models for every training point is generally infeasible. Instead, the influence function relies on a heavier approximation to estimate the effect of perturbing a single training point, rather than actually forgetting the training samples. In contrast, our approach only requires running the unlearning algorithm \textit{once} for a synthesized image query. This allows us to use a milder approximation and obtain a model that forgets the synthesized sample.
\arxiv{Guo~\etal~\cite{guo2019certified} and Tanno~\etal~\cite{tanno2022repairing} explore a similar formulation for unlearning training images and draw a connection between their unlearning algorithms and influence function. Our approach aims to unlearn the synthesized image instead, which connects to influence function in a similar fashion. We discuss our method's connection to influence function in more detail in Appendix~\ref{sec:supp_connect_infl}.}

\myparagraph{Optimizing a subset of weights.}
To further regularize the unlearning process, we optimize a small subset of weights, specifically $W^k$ and $W^v$, the key and value projection matrices in cross-attention layers~\cite{bahdanau2015neural,vaswani2017attention}. In text-to-image models, cross-attention facilitates text-to-image binding, where $W^k$ identifies which features match each text token, while $W^v$ determines how to modify the features for the matched patches. We observe that performing unlearning $W^k$ and $W^v$ is effective for attribution. Prior works also select the same set of parameters to improve fine-tuning~\cite{kumari2022customdiffusion} and unlearning~\cite{kumari2023conceptablation}.

\myparagraph{Implementation details.}
We conduct our studies on text-conditioned latent diffusion models~\cite{rombach2021highresolution}. %
Since diffusion models are typically trained with $T=1000$ steps, evaluating the loss for all timesteps is costly. Therefore, we speed up computation by calculating the loss $\mathcal{L}(\mathbf{z}, \theta)$ with strided timesteps; we find that using a stride of 50 or 100 leads to good attribution performance. For calculating the loss change $\Delta \mathcal{L}(\hat{\mathbf{z}}, \theta)$ during evaluation, we take a finer stride of 5 steps to ensure a more accurate estimation of the DDPM loss. Additional details of our method, including hyperparameter choices, are provided in Appendix~\ref{sec:supp_impl}.

\section{Experiments}
\label{sec:experiment}
\arxiv{We validate our method in two ways. The first is a reliable, ``gold-standard'', but intensive -- retraining a model from scratch without influential images identified by the algorithm. In Section~\ref{sec:exp_leave_k_out}, we perform this evaluation on a medium-sized dataset of 100k MSCOCO images~\cite{lin2014microsoft}. Secondly, in Section~\ref{sec:exp_abc}, we evaluate our method on the Customized Model Benchmark~\cite{wang2023evaluating}, which measures attribution through customization on Stable Diffusion models~\cite{rombach2021highresolution}. This tests how well our method can apply to large-scale text-to-image models.}

\begin{figure}
    \centering
    \includegraphics[width=\linewidth]{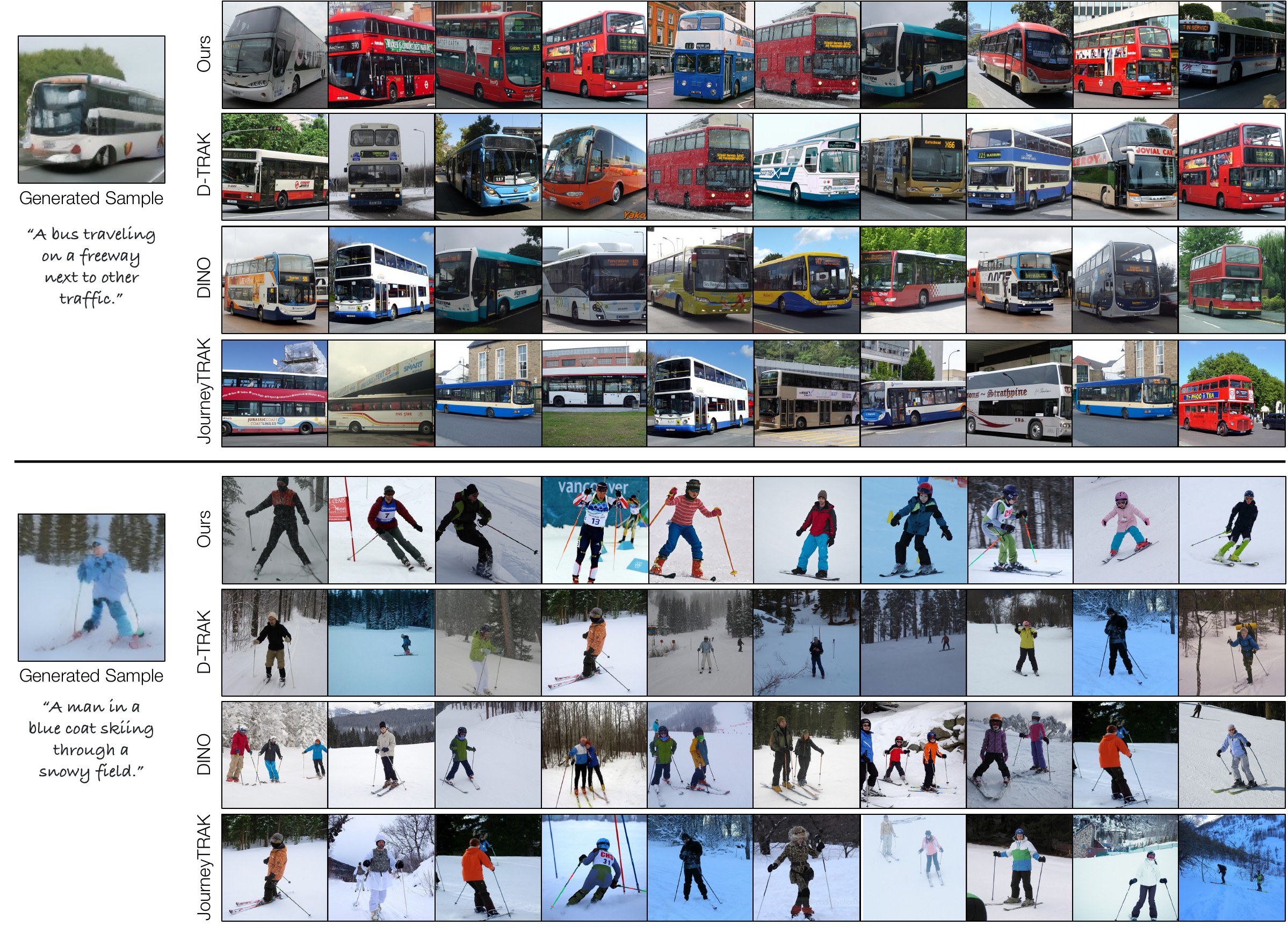}
    \vspace{-6mm}
    \caption{\textbf{Attribution results on MSCOCO models.} We show generated samples used as a query on the left, with training images being identified by different methods on the right. Qualitatively, our method retrieves images with more similar visual attributes. Notably, our method better matches the poses of the buses (considering random flips during training) and the poses and enumeration of skiers.}
    \vspace{-2mm}
    \label{fig:qual_coco_baseline}
\end{figure}

\subsection{Leave-$K$-out counterfactual evaluation}
\label{sec:exp_leave_k_out}
\myparagraph{Evaluation protocol.}
We select latent diffusion models~\cite{rombach2021highresolution} trained on MSCOCO~\cite{lin2014microsoft}, as its moderate size (118,287 images) allows for repeated leave-$K$-out retraining. Specifically, we use the pre-trained model evaluated in Georgiev~\etal~\cite{georgiev2023journey}. As outlined in Section~\ref{sec:method_eval}, for each synthesized image $\hat{\mathbf{z}}$, we measure the leave-$K$-out model's \textbf{(1) loss change} $\Delta \mathcal{L}(\hat{\mathbf{z}}, \theta)$ and \textbf{(2) deviation of generation} $\Delta G_\theta(\epsilon, \mathbf{c})$. The deviation is measured by mean square error (MSE) and CLIP similarity~\cite{radford2021learning}.
\arxiv{We collect 110 synthesized images from the pre-trained model for evaluation, with different text prompts sourced from the MSCOCO validation set. We evaluate $\Delta \mathcal{L}(\hat{\mathbf{z}}, \theta)$ and $\Delta G_\theta(\epsilon, \mathbf{c})$ for all synthesized images and report mean and standard error.}

We compare our method with several baselines: 
\vspace{-1mm}
\vspace{-\topsep}
\begin{itemize}[leftmargin=12pt]
  \setlength{\parskip}{0pt}
  \setlength{\itemsep}{-8pt}
    \item \textbf{Random:} \camready{We train models with $K$ random images removed, using 10 models per value of $K$.} \\
    \item \textbf{Image similarity:} pixel space, CLIP image features~\cite{radford2021learning}, DINO~\cite{caron2021emerging}, and DINOv2~\cite{oquab2023dinov2} \\
    \item \textbf{Text similarity:} CLIP text features \\
    \item \textbf{Attribution by Customization~\cite{wang2023evaluating} (AbC):} fine-tuned image features trained on the Customized Model benchmark, denoted as CLIP (AbC) and DINO (AbC) \\
    \item \camready{\textbf{DataInf~\cite{kwon2024datainf}}: an influence estimation method based on approximating matrix inverses.} \\
    \item \camready{\textbf{TRAK~\cite{park2023trak} and JourneyTRAK~\cite{georgiev2023journey}} are influence function-based methods that match the loss gradients of training and synthesized images, using random projection for efficiency. Both methods run the influence function on multiple models trained on the same dataset (20 in this test) and average the scores. The main difference is in the diffusion loss calculation: TRAK randomly samples and averages the loss over timesteps, while JourneyTRAK calculates it only at $t=400$ for synthesized images during counterfactual evaluation.\\}
    \item \camready{\textbf{D-TRAK~\cite{zheng2023intriguing}:} a concurrent work that extends TRAK by changing the denoising loss function into a square loss during influence computation. As mentioned by the authors, D-TRAK yields unexpectedly stronger performance even though the design choice is not theoretically understood. Different from TRAK, D-TRAK yields competitive performance with a single model.}

\end{itemize}
\vspace{-\topsep}
\arxiv{For attribution methods, we use $K = 500, 1000,$ and $4000$, representing approximately $0.42\%$, $0.85\%$, and $3.4\%$ of the MSCOCO dataset, respectively.}

\myparagraph{Visual comparison of attributed images.}
In Figure~\ref{fig:qual_coco_baseline}, we find that our method, along with other baselines, can attribute synthesized images to visually similar training images. However, our method more consistently attributes images with the same fine-grained attributes, such as object location, pose, and counts. We provide more results in Appendix~\ref{sec:supp_mscoco_additional_analysis}. Next, we proceed with the counterfactual analysis, where we test whether these attributed images are truly \textit{influential}.

\begin{table}[]
{\small
\centering
\resizebox{1.\linewidth}{!}{
\begin{tabular}{llccccccccc}
\toprule
\multirow{2}{*}{Family} & \multirow{2}{*}{Method} & \multicolumn{3}{c}{$\Delta \mathcal{L}(\hat{\mathbf{z}}, \theta)$ $\uparrow$ {\footnotesize(x$10^{-3}$)}} & \multicolumn{3}{c}{$\Delta G_\theta(\epsilon, \mathbf{c})$ (MSE) $\uparrow$ {\footnotesize(x$10^{-2}$)}} & \multicolumn{3}{c}{$\Delta G_\theta(\epsilon, \mathbf{c})$ (CLIP) $\downarrow$ {\footnotesize(x$10^{-1}$)}} \\ \cmidrule(lr){3-5}\cmidrule(lr){6-8}\cmidrule(lr){9-11} 
 &  & \multicolumn{1}{c}{K=500} & \multicolumn{1}{c}{K=1000} & \multicolumn{1}{c}{K=4000} & \multicolumn{1}{c}{K=500} & \multicolumn{1}{c}{K=1000} & \multicolumn{1}{c}{K=4000} & \multicolumn{1}{c}{K=500} & \multicolumn{1}{c}{K=1000} & \multicolumn{1}{c}{K=4000} \\ \midrule
Random & Random & 3.5{\color{gray}\scriptsize±0.03} & 3.5{\color{gray}\scriptsize±0.03} & 3.5{\color{gray}\scriptsize±0.03} & 4.1{\color{gray}\scriptsize±0.06} & 4.1{\color{gray}\scriptsize±0.06} & 4.0{\color{gray}\scriptsize±0.06} & 7.9{\color{gray}\scriptsize±0.03} & 7.8{\color{gray}\scriptsize±0.03} & 7.9{\color{gray}\scriptsize±0.03} \\ \hdashline[0.5pt/1pt]
Pixel & Pixel & 3.6{\color{gray}\scriptsize±0.10} & 3.6{\color{gray}\scriptsize±0.10} & 4.0{\color{gray}\scriptsize±0.11} & 4.3{\color{gray}\scriptsize±0.19} & 4.3{\color{gray}\scriptsize±0.21} & 4.9{\color{gray}\scriptsize±0.21} & 7.9{\color{gray}\scriptsize±0.10} & 7.8{\color{gray}\scriptsize±0.09} & 7.7{\color{gray}\scriptsize±0.10} \\ \hdashline[0.5pt/1pt]
Text & CLIP Text & 3.8{\color{gray}\scriptsize±0.12} & 4.2{\color{gray}\scriptsize±0.14} & 5.5{\color{gray}\scriptsize±0.25} & 4.1{\color{gray}\scriptsize±0.20} & 4.3{\color{gray}\scriptsize±0.19} & 4.6{\color{gray}\scriptsize±0.19} & 7.8{\color{gray}\scriptsize±0.08} & 7.7{\color{gray}\scriptsize±0.09} & 7.4{\color{gray}\scriptsize±0.08} \\ \hdashline[0.5pt/1pt]
\multirow{3}{*}{Image} & DINOv2 & 3.9{\color{gray}\scriptsize±0.12} & 4.3{\color{gray}\scriptsize±0.15} & 6.3{\color{gray}\scriptsize±0.33} & 4.3{\color{gray}\scriptsize±0.20} & 4.6{\color{gray}\scriptsize±0.20} & 5.1{\color{gray}\scriptsize±0.19} & 7.7{\color{gray}\scriptsize±0.09} & 7.7{\color{gray}\scriptsize±0.08} & 7.0{\color{gray}\scriptsize±0.09} \\
 & CLIP & 4.2{\color{gray}\scriptsize±0.14} & 4.7{\color{gray}\scriptsize±0.18} & 6.4{\color{gray}\scriptsize±0.32} & 4.4{\color{gray}\scriptsize±0.20} & 4.6{\color{gray}\scriptsize±0.21} & 5.2{\color{gray}\scriptsize±0.22} & 7.6{\color{gray}\scriptsize±0.09} & 7.5{\color{gray}\scriptsize±0.08} & 6.8{\color{gray}\scriptsize±0.08} \\
 & DINO & 4.8{\color{gray}\scriptsize±0.15} & 5.6{\color{gray}\scriptsize±0.20} & 8.1{\color{gray}\scriptsize±0.35} & 4.5{\color{gray}\scriptsize±0.16} & 5.3{\color{gray}\scriptsize±0.22} & 5.9{\color{gray}\scriptsize±0.21} & \underline{7.4{\color{gray}\scriptsize±0.09}} & \underline{7.1{\color{gray}\scriptsize±0.09}} & \textbf{6.3{\color{gray}\scriptsize±0.10}} \\ \hdashline[0.5pt/1pt]
\multirow{2}{*}{AbC} & CLIP (AbC) & 4.4{\color{gray}\scriptsize±0.13} & 4.9{\color{gray}\scriptsize±0.17} & 6.9{\color{gray}\scriptsize±0.32} & 4.6{\color{gray}\scriptsize±0.20} & 5.0{\color{gray}\scriptsize±0.22} & 5.6{\color{gray}\scriptsize±0.23} & 7.5{\color{gray}\scriptsize±0.09} & 7.3{\color{gray}\scriptsize±0.09} & 6.5{\color{gray}\scriptsize±0.09} \\
 & DINO (AbC) & 4.8{\color{gray}\scriptsize±0.15} & 5.5{\color{gray}\scriptsize±0.20} & 8.1{\color{gray}\scriptsize±0.35} & 4.8{\color{gray}\scriptsize±0.22} & 4.9{\color{gray}\scriptsize±0.20} & 5.8{\color{gray}\scriptsize±0.21} & 7.5{\color{gray}\scriptsize±0.09} & 7.2{\color{gray}\scriptsize±0.09} & \textbf{6.3{\color{gray}\scriptsize±0.09}} \\ \hdashline[0.5pt/1pt]
\multirow{4}{*}{Influence} & DataInf & 3.7{\color{gray}\scriptsize±0.11} & 3.7{\color{gray}\scriptsize±0.12} & 3.9{\color{gray}\scriptsize±0.13} & 4.1{\color{gray}\scriptsize±0.18} & 4.1{\color{gray}\scriptsize±0.20} & 4.2{\color{gray}\scriptsize±0.18} & 7.9{\color{gray}\scriptsize±0.08} & 7.9{\color{gray}\scriptsize±0.09} & 7.8{\color{gray}\scriptsize±0.10} \\
 & TRAK & 5.2{\color{gray}\scriptsize±0.14} & 5.8{\color{gray}\scriptsize±0.16} & 7.1{\color{gray}\scriptsize±0.16} & 4.7{\color{gray}\scriptsize±0.21} & 4.7{\color{gray}\scriptsize±0.24} & 4.7{\color{gray}\scriptsize±0.20} & 7.6{\color{gray}\scriptsize±0.09} & 7.6{\color{gray}\scriptsize±0.09} & 7.5{\color{gray}\scriptsize±0.09} \\
 & JourneyTRAK & 4.4{\color{gray}\scriptsize±0.11} & 4.9{\color{gray}\scriptsize±0.13} & 5.7{\color{gray}\scriptsize±0.15} & 4.8{\color{gray}\scriptsize±0.19} & 5.4{\color{gray}\scriptsize±0.23} & 5.4{\color{gray}\scriptsize±0.24} & 7.7{\color{gray}\scriptsize±0.09} & 7.5{\color{gray}\scriptsize±0.09} & 7.5{\color{gray}\scriptsize±0.09} \\ 
 & D-TRAK & \underline{5.4{\color{gray}\scriptsize±0.16}} & \underline{6.6{\color{gray}\scriptsize±0.22}} & \underline{9.6{\color{gray}\scriptsize±0.33}} & \textbf{5.9{\color{gray}\scriptsize±0.24}} & \textbf{6.4{\color{gray}\scriptsize±0.25}} & \textbf{7.8{\color{gray}\scriptsize±0.30}} & \textbf{7.3{\color{gray}\scriptsize±0.10}} & \underline{7.1{\color{gray}\scriptsize±0.09}} & \underline{6.4{\color{gray}\scriptsize±0.09}} \\ \hdashline[0.5pt/1pt]
Ours & Ours & \textbf{5.6{\color{gray}\scriptsize±0.16}} & \textbf{6.7{\color{gray}\scriptsize±0.22}} & \textbf{9.8{\color{gray}\scriptsize±0.32}} & \underline{5.1{\color{gray}\scriptsize±0.21}} & \underline{5.7{\color{gray}\scriptsize±0.24}} & \underline{6.1{\color{gray}\scriptsize±0.22}} & \textbf{7.3{\color{gray}\scriptsize±0.09}} & \textbf{7.0{\color{gray}\scriptsize±0.09}} & \underline{6.4{\color{gray}\scriptsize±0.10}} \\ \bottomrule
\end{tabular}}}
\vspace{1pt}
\caption{\textbf{Leave-$K$-out baseline comparisons.} \camready{Given a synthesized image  $\hat{\mathbf{z}}$, we train leave-$K$-out models for each of the attribution methods and track $\Delta \mathcal{L}(\hat{\mathbf{z}}, \theta)$, the increase in loss change, and $\Delta G_\theta(\epsilon, \mathbf{c})$, deviation of generation. We report results over 110 samples, and {\color{gray} gray} shows the standard error. \textbf{Bolded} and \underline{underlined} are the best and second best performing method, respectively.}}
\label{tab:big_results}
\end{table}

\myparagraph{Tracking loss changes in leave-$K$-out models.}
First, we report the change in DDPM loss for leave-$K$-out models in %
\camready{Table~\ref{tab:big_results}}.
Matching in plain pixel or text feature space yields weak performance, while deep image features, particularly DINO, perform better. Interestingly, DINO outperforms \camready{most} influence function methods at $K=4000$, despite not being trained specifically for the attribution task. Fine-tuning image features with the Customized Model benchmark, such as CLIP (AbC), shows some improvement. However, in general, the improvement is limited, indicating that transferring from attributing customized models to general models remains challenging~\cite{wang2023evaluating}.

Among influence functions, \camready{DataInf performs poorly. According to the paper, the matrix inverse approximation DataInf uses is more suitable for LoRA fine-tuned models rather than text-to-image models trained from scratch. This approximation leads to poor performance.}

TRAK significantly outperforms JourneyTRAK. We hypothesize that this is because JourneyTRAK collects gradients only for denoising loss at timestep $t=400$, making it less effective for identifying influential images that affect the DDPM training loss across different noise levels. \camready{On the other hand, D-TRAK is the best performing influence-function-based method, although the reason behind its strong performance is not well-understood~\cite{zheng2023intriguing}.}

Our method consistently performs best across all $K$ values, outperforming both influence functions and feature-matching methods. %

\begin{figure}
    \centering
    \includegraphics[width=\linewidth]{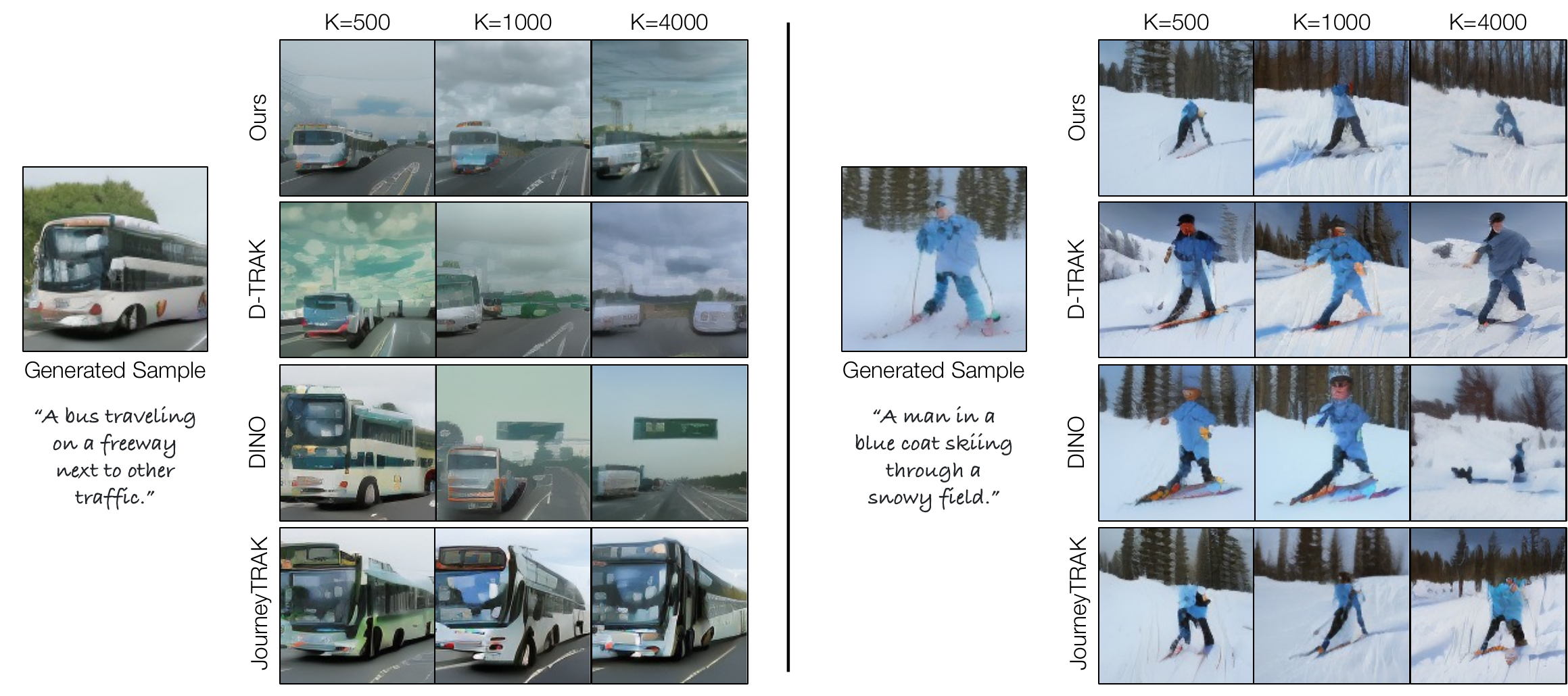}
    \caption{\textbf{Leave-$K$-out analysis for MSCOCO models.} We compare images across our method and baselines generated by leave-$K$-out models, using different $K$ values, all under the same random noise and text prompt. A significant deviation in regeneration indicates that critical, influential images were identified by the attribution algorithm. \camready{Our method leads to image generation that deviate significantly, even with as few as 500 influential images removed ($\sim$0.42$\%$ of the dataset).}}
    \vspace{-2mm}
    \label{fig:qual_coco_leave_k}
\end{figure}

\myparagraph{Deviation of generated output in leave-$K$-out models.}
Figure~\ref{fig:qual_coco_leave_k} \camready{and Table~\ref{tab:big_results}} shows the deviation in generated outputs for leave-$K$-out models, where all images are generated using the same noise input and text prompt. \camready{Consistent with the loss change evaluation, D-TRAK, DINO, and our method yield the largest deviation with a small budget. While the three methods perform similarly well in CLIP similarity, D-TRAK outperforms our method in MSE deviations. In contrast, TRAK and JourneyTRAK have formulations similar to D-TRAK, but they perform subpar in this test. Interestingly, while D-TRAK yields significant performance gain by changing the loss function, we did not observe the same improvement when applying the same loss function changes to our method. We discuss this more in Appendix~\ref{sec:supp_mscoco_additional_analysis}.}.

\camready{In addition, we include analysis on whether unlearned models and leave-$K$-out models forget only the specific concept, along with more qualitative results, in Appendix~\ref{sec:supp_mscoco_additional_analysis}.}

\myparagraph{Ablation studies.}
\camready{We study two design choices:  (1) the effect of EWC regularization and (2) different weight subsets for optimization. Table~\ref{tab:ablation_studies} shows the results. We find that unlearning without EWC loss greatly hurts attribution performance, indicating the importance of regulating unlearning with Fisher information. We also find that using a subset of weights to unlearn leads to better attribution in general. We test three weight subset selection schemes (Attn, Cross Attn, Cross Attn KV), all of which outperform the version using all weights. Among them, updating Cross Attn KV performs the best, consistent with findings from model customization~\cite{kumari2022customdiffusion,tewel2023keylocked} and unlearning~\cite{kumari2023conceptablation}.}

\begin{figure}
    \centering
    \includegraphics[width=\linewidth]{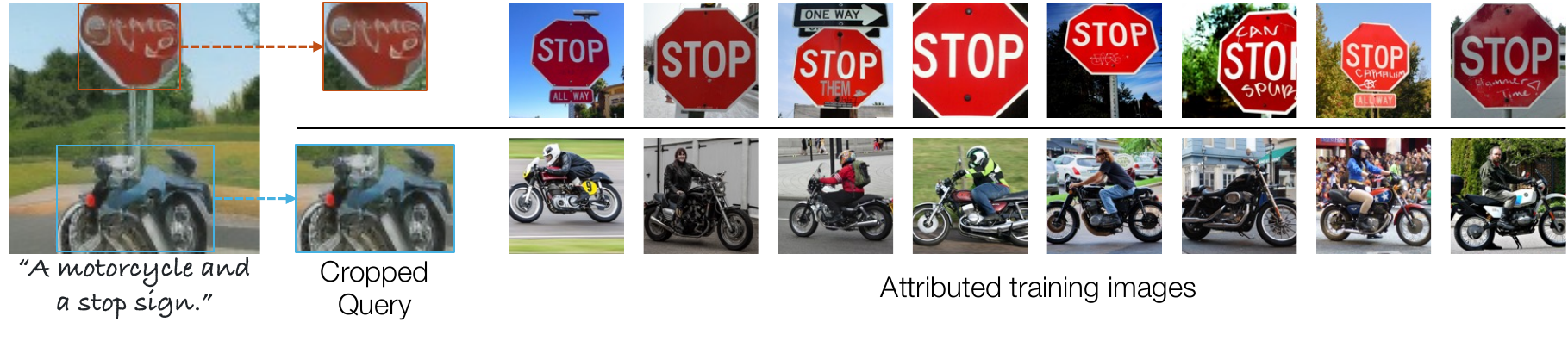}
    \vspace{-8mm}
    \caption{\textbf{Spatially-localized attribution.} \arxiv{Given a synthesized image (left), we crop regions containing specific objects using GroundingDINO~\cite{liu2023grounding}. We attribute each object separately by only running forgetting on the pixels within the cropped region. Our method can attribute different synthesized regions to different training images.}}
    \vspace{-2mm}
    \label{fig:coco_comp}
\end{figure}

\begin{table}[]
{\small
\centering
\resizebox{1.\linewidth}{!}{
\begin{tabular}{lcccccccccc}
\toprule
 &  & \multicolumn{3}{c}{$\Delta \mathcal{L}(\hat{\mathbf{z}}, \theta)$ $\uparrow$ {\footnotesize(x$10^{-3}$)}} & \multicolumn{3}{c}{$\Delta G_\theta(\epsilon, \mathbf{c})$ (MSE) $\uparrow$ {\footnotesize(x$10^{-2}$)}} & \multicolumn{3}{c}{$\Delta G_\theta(\epsilon, \mathbf{c})$ (CLIP) $\downarrow$ {\footnotesize(x$10^{-1}$)}} \\ \cmidrule(lr){3-5}\cmidrule(lr){6-8}\cmidrule(lr){9-11}
\multirow{-2}{*}{Variation} & \multirow{-2}{*}{EWC?} & \multicolumn{1}{c}{K=500} & \multicolumn{1}{c}{K=1000} & \multicolumn{1}{c}{K=4000} & \multicolumn{1}{c}{K=500} & \multicolumn{1}{c}{K=1000} & \multicolumn{1}{c}{K=4000} & \multicolumn{1}{c}{K=500} & \multicolumn{1}{c}{K=1000} & \multicolumn{1}{c}{K=4000} \\ \midrule
SGD (1 step) & & 3.5{\color{gray}\scriptsize±0.10} & 3.5{\color{gray}\scriptsize±0.10} & 3.5{\color{gray}\scriptsize±0.11} & 3.9{\color{gray}\scriptsize±0.16} & 3.9{\color{gray}\scriptsize±0.17} & 4.1{\color{gray}\scriptsize±0.20} & 7.9{\color{gray}\scriptsize±0.09} & 7.8{\color{gray}\scriptsize±0.09} & 7.8{\color{gray}\scriptsize±0.09} \\
SGD (10 step) & & 3.4{\color{gray}\scriptsize±0.10} & 3.5{\color{gray}\scriptsize±0.10} & 3.5{\color{gray}\scriptsize±0.10} & 3.7{\color{gray}\scriptsize±0.16} & 3.9{\color{gray}\scriptsize±0.17} & 3.9{\color{gray}\scriptsize±0.15} & 7.9{\color{gray}\scriptsize±0.09} & 7.9{\color{gray}\scriptsize±0.09} & 7.8{\color{gray}\scriptsize±0.09} \\ \hdashline[0.5pt/1pt]
Full & \checkmark & 5.5{\color{gray}\scriptsize±0.17} & 6.2{\color{gray}\scriptsize±0.20} & 8.4{\color{gray}\scriptsize±0.27} & 4.6{\color{gray}\scriptsize±0.19} & 5.2{\color{gray}\scriptsize±0.23} & 5.5{\color{gray}\scriptsize±0.23} & 7.6{\color{gray}\scriptsize±0.08} & 7.4{\color{gray}\scriptsize±0.09} & 7.1{\color{gray}\scriptsize±0.10} \\
Attn & \checkmark & 5.5{\color{gray}\scriptsize±0.17} & 6.6{\color{gray}\scriptsize±0.23} & 9.3{\color{gray}\scriptsize±0.33} & \textbf{5.1{\color{gray}\scriptsize±0.23}} & 5.3{\color{gray}\scriptsize±0.22} & \textbf{6.2{\color{gray}\scriptsize±0.22}} & 7.4{\color{gray}\scriptsize±0.10} & 7.2{\color{gray}\scriptsize±0.09} & 6.7{\color{gray}\scriptsize±0.09} \\
Cross Attn & \checkmark & 5.5{\color{gray}\scriptsize±0.17} & 6.5{\color{gray}\scriptsize±0.24} & 9.1{\color{gray}\scriptsize±0.33} & 4.8{\color{gray}\scriptsize±0.20} & 5.3{\color{gray}\scriptsize±0.22} & 6.0{\color{gray}\scriptsize±0.23} & \textbf{7.3{\color{gray}\scriptsize±0.10}} & 7.2{\color{gray}\scriptsize±0.10} & 6.6{\color{gray}\scriptsize±0.09} \\
Cross Attn KV & \checkmark & \textbf{5.6{\color{gray}\scriptsize±0.16}} & \textbf{6.7{\color{gray}\scriptsize±0.22}} & \textbf{9.8{\color{gray}\scriptsize±0.32}} & \textbf{5.1{\color{gray}\scriptsize±0.21}} & \textbf{5.7{\color{gray}\scriptsize±0.24}} & 6.1{\color{gray}\scriptsize±0.22} & \textbf{7.3{\color{gray}\scriptsize±0.09}} & \textbf{7.0{\color{gray}\scriptsize±0.09}} & \textbf{6.4{\color{gray}\scriptsize±0.10}} \\ \bottomrule
\end{tabular}}}
\vspace{1pt}
\caption{\textbf{Leave-$K$-out ablation studies.} \camready{We ablate our design choices and report $\Delta \mathcal{L}(\hat{\mathbf{z}}, \theta)$ and $\Delta G_\theta(\epsilon, \mathbf{c})$ as in Table~\ref{tab:big_results}. We find that naive unlearning (SGD without EWC regularization) is not effective. We then compare four different sets of weights to unlearn and find that cross-attention $W^k$, $W^v$ (Cross Attn KV) outperforms other configurations. \textbf{Bolded} are the best performing method, and {\color{gray} gray} shows the standard error.}}
\label{tab:ablation_studies}
\end{table} 
\myparagraph{Spatially-localized attribution.}
\arxiv{
While our formulation is written for whole images, we can run attribution on specific regions with little modification.
We demonstrate this in Figure~\ref{fig:coco_comp} on a generated image of a motorcycle and stop sign, using bounding boxes identified by GroundingDINO~\cite{liu2023grounding}. For each detected object, we run our unlearning (using the same prompt) on that specific object by optimizing the objective only within the bounding box. By doing so, we attribute different training images for the stop sign and motorcycle.}

\subsection{Customized Model Benchmark}
\label{sec:exp_abc}
\arxiv{Wang~\etal~\cite{wang2023evaluating} focus on a specialized form of attribution: attributing customized models trained on an individual or a few exemplar images. This approach provides ground truth attribution, since the images generated by customized models are computationally influenced by exemplar images. While this evaluation has limited generalization to attribution performance with larger training sets, it is the only tractable evaluation for attributing large-scale text-to-image models to date.}

\myparagraph{Evaluation protocol.}
\arxiv{Since the Customized Model Benchmark has ground truth, the problem is evaluated as a retrieval task. We report \textbf{Recall@K} and \textbf{mAP}, measuring the success of retrieving the exemplar images amongst a set including 100K LAION images. We compare with Wang~\etal's feature-matching approach that finetunes on the Customized Model dataset, referred to as DINO (AbC) and CLIP (AbC).
\camready{We also compare with D-TRAK, the best-performing influence function method in our previous MSCOCO experiment.} For our evaluation, we selected a subset of the dataset comprising 20 models: 10 object-centric and 10 artist-style models. We select 20 synthesized images with different prompts for each model, resulting in 400 synthesized image queries.}

\begin{figure}
    \centering
    \includegraphics[width=\linewidth]{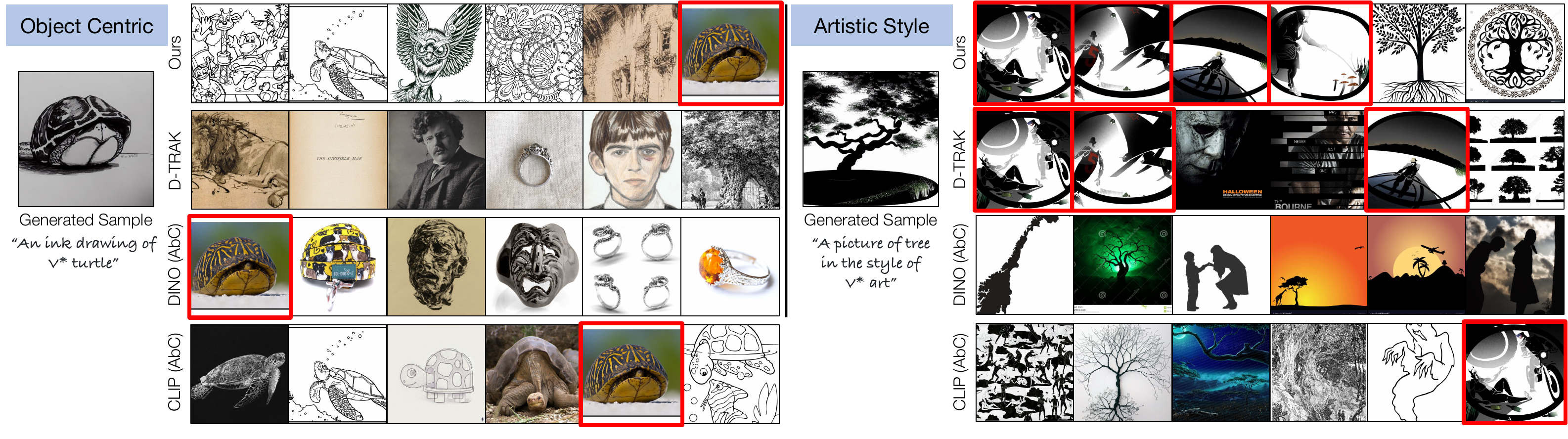}
    \vspace{-6mm}
    \caption{\textbf{Qualitative examples on the Customized Model benchmark.} The red boxes indicate ground truth exemplar images used for customizing the model. Both our method and \camready{AbC} baselines successfully identify the exemplar images on object-centric models (left), while our method outperforms the baselines with artistic style models (right).
    }
    \label{fig:qual_abc_baseline}
\end{figure}

\begin{figure}
    \centering
    \begin{tabular}{cc}
        \hspace{-2mm}
        \includegraphics[width=0.48\linewidth]{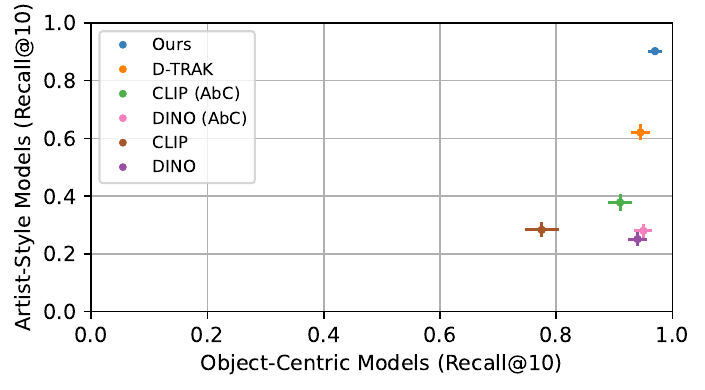} & 
        \includegraphics[width=0.48\linewidth]{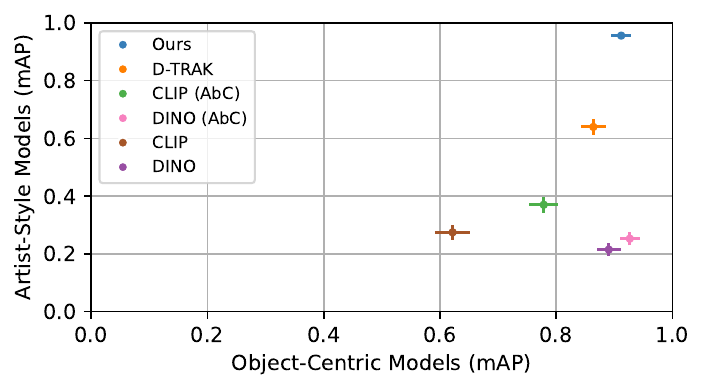} 
    \end{tabular}
    \vspace{-4mm}
    \caption{\textbf{Customized Model benchmark~\cite{wang2023evaluating}.} We report Recall@10 (left) and mAP (right) and show performance on artist-style models (y-axis) vs. object-centric models (x-axis). On object-centric models, our method performs on par with AbC features, which were directly tuned on the benchmark, while significantly outperforming them on artist-style models. \camready{D-TRAK performs the second best on artist-style models but worse on object-centric models.} We plot one standard error on both axes.}
    \label{fig:abc_baseline}
\end{figure}

\myparagraph{Comparing with \camready{other methods}.}
We report Recall@10 and mAP in Figure~\ref{fig:abc_baseline}. Our method performs on par with baselines for object-centric models, while significantly outperforming them on artist-style models. Although CLIP (AbC) and DINO (AbC) are fine-tuned for this attribution task, the feature matching approach can sometimes confuse whether to attribute a synthesized image to style-related or object-related images. In contrast, our method, which has access to the model itself, traces influential training images more effectively. \camready{D-TRAK performs worse than our method despite also having model access, which suggests that the influence approximations could be less accurate for larger models.} In Figure~\ref{fig:qual_abc_baseline}, we show a qualitative example. While DINO (AbC) and CLIP (AbC) can retrieve visually or semantically similar images, our method successfully identifies the exemplars in both cases. We include ablation studies in Appendix~\ref{sec:supp_abc_additional_analysis}.

\section{Discussion, Broader Impacts, and Limitations}
\label{sec:discussion}

Generative models have entered the public consciousness, spawning companies and ecosystems that are deeply impacting the creative industry. The technology raises high-stakes ethical and legal questions surrounding the authorship of generated content~\cite{lee2023talkin,elkin2023can,hacohen2024not,asogai}. Data attribution is a critical piece of understanding the behavior of generative models, with potential applications in informing a compensation model for rewarding contributors for training data. In addition, data attribution can join other works~\cite{bau2019gandissect,meng2022locating,foote2023neuron,dravid2023rosetta} as a set of tools that allow end users to interpret how and why a model behaves, enabling a more trustworthy environment for machine learning models.

Our work proposes a method for data attribution for text-to-image models, leveraging model unlearning. We provide a counterfactual validation, verifying that removing the identified influential images indeed destroys the target image. While our method empirically demonstrates that unlearning can be effectively used, work remains to make this practical. Though our model unlearns efficiently, estimating the reconstruction loss on the training set remains a bottleneck, as several forward passes are required on each training estimate, \camready{as profiled in Appendix~\ref{sec:supp_runtime}}. \arxiv{While our evaluation showed that unlearning is useful for attribution, direct evaluation of unlearning algorithms for large generative models remains an open research challenge.} Furthermore, to find a critical set of images, our method and baselines assign influence scores to individual images and sort them. However, groups of images may have interactions that are not captured in such a system. Furthermore, our method and baselines explore attribution of the whole image, while finer attribution on individual aspects of the image, such as style, structure, or individual segments, are of further interest.

\myparagraph{Acknowledgements.}
We thank Kristian Georgiev for answering our inquiries regarding JourneyTRAK implementation and evaluation and providing us with their models and an earlier version of the JourneyTRAK code. We thank Nupur Kumari, Kangle Deng, and Grace Su for their feedback on the draft. This work is partly supported by the Packard Fellowship, JPMC Faculty Research Award, NSF
IIS-2239076, and NSF IIS-2403303.  %

\medskip

{
\small
\bibliographystyle{unsrt}
\bibliography{main}
}

\appendix

\clearpage

\section{Derivations for Unlearning}
\label{sec:supp_unlearning_math}
In Section~\ref{sec:methods}, we describe our unlearning method and its relationship to influence functions. Here, we provide more detailed derivations. Let $\theta_0$ be the pretrained model trained on dataset $\mathcal{D}$ and loss $\sum_{\mathbf{z}\in \mathcal{D}} \mathcal{L}(\mathbf{z}, \theta)$. $N$ is the size of the dataset. Our goal is to obtain a model $\theta_{-\hat{\mathbf{z}}}$ that unlearns the synthesized image $\hat{\mathbf{z}}$.

\subsection{EWC Loss}
We summarize EWC Loss~\cite{kirkpatrick2017overcoming}, which is the second order Taylor approximation of the data loss $\sum_{\mathbf{z}\in \mathcal{D}} \mathcal{L}(\mathbf{z}, \theta)$ around $\theta_0$. We denote the Hessian of the loss as $H_{\theta_0}$, where $\left[H_{\theta_0}\right]_{ij} = \frac1N\frac{\partial^2}{\partial\theta_i\partial\theta_j} \sum_{\mathbf{z}\in \mathcal{D}} \mathcal{L}(\mathbf{z}, \theta)|_{\theta=\theta_0}$.
We denote the remainder term as $R(\theta)$.

\begin{equation}
\begin{aligned}
    \sum_{\mathbf{z}\in \mathcal{D}} \mathcal{L}(\mathbf{z}, \theta) =& \sum_{\mathbf{z}\in \mathcal{D}} \mathcal{L}(\mathbf{z}, \theta_0) + \sum_{\mathbf{z}\in \mathcal{D}} \nabla\mathcal{L}(\mathbf{z}, \theta)|_{\theta=\theta_0} (\theta-\theta_0) + \frac{N}{2}(\theta-\theta_0)^TH_{\theta_0} (\theta-\theta_0) + R(\theta) \\
    &\approx \sum_{\mathbf{z}\in \mathcal{D}} \mathcal{L}(\mathbf{z}, \theta_0) + \frac{N}{2}(\theta-\theta_0)^TH_{\theta_0} (\theta-\theta_0).
\end{aligned}
\end{equation}

We assume that pretrained $\theta_0$ is near the local minimum of the loss, resulting in a near-zero gradient $\sum_{\mathbf{z}\in \mathcal{D}} \nabla\mathcal{L}(\mathbf{z}, \theta_0)|_{\theta=\theta_0}$. We drop out higher order remainder term $R(\theta)$ in our approximation. If the model is trained with a negative log-likelihood loss, the Hessian $H_{\theta_0}$ is equivalent to the Fisher information $F$~\cite{grosse2023studying}, leading to:

\begin{equation}
\begin{aligned}
    \sum_{\mathbf{z}\in \mathcal{D}} \mathcal{L}(\mathbf{z}, \theta) &\approx  \sum_{\mathbf{z}\in \mathcal{D}}\mathcal{L}(\mathbf{z}, \theta_0) + \frac{N}{2}(\theta-\theta_0)^TF (\theta-\theta_0).
\end{aligned}
\end{equation}

We note that we focus on diffusion models, which are trained on a lower bound of the log-likelihood. \arxiv{In this context, the Fisher information can be viewed as the Gauss-Newton approximation of the Hessian~\cite{martens2020new}.} The formulation satisfies the purpose of approximating the training loss on the dataset and serves as an effective regularizer on our unlearning objective. 

\myparagraph{Diagonal approximation of Fisher.}
\arxiv{Since Fisher information is the covariance of the log-likelihood gradient, its diagonal approximation is equivalent to taking the square of the gradients and averaging them across the training set. This diagonal approximation is adopted by EWC loss~\cite{kirkpatrick2017overcoming,li2020fewshot}. In the context of diffusion models, the Fisher information is estimated by averaging across training data, random noise, and timesteps.}

\subsection{Updating step for unlearning}

We then derive the Newton update of our unlearning objective in Equation~\ref{eqn:newton_removal}. Below, we repeat our unlearning objective in Equation~\ref{eqn:unlearning}:

\begin{equation}
\begin{aligned}
    \mathcal{L}_\text{unlearn}^{\hat{\mathbf{z}}}(\theta) 
    &\approx -\mathcal{L}(\hat{\mathbf{z}}, \theta) + \frac{N}{2}(\theta-\theta_0)^TF (\theta-\theta_0).
\end{aligned}
\end{equation}

The Newton step is a second-order update of the form below, where $\alpha$ controls the step size:

\begin{equation}
\begin{aligned}
    \theta \leftarrow \theta - \alpha \left[H_{\text{unlearn}}^{\hat{\textbf{z}}}(\theta) \right]^{-1} \nabla \mathcal{L}_\text{unlearn}^{\hat{\mathbf{z}}}(\theta),
\end{aligned}
\label{eqn:supp_newton}
\end{equation}
where $H_{\text{unlearn}}^{\hat{\textbf{z}}}(\theta)$ is the Hessian of $\mathcal{L}_\text{unlearn}^{\hat{\mathbf{z}}}(\theta)$. Now we derive $\nabla \mathcal{L}_\text{unlearn}^{\hat{\mathbf{z}}}(\theta)$:

\begin{equation}
\begin{aligned}
    \nabla \mathcal{L}_\text{unlearn}^{\hat{\mathbf{z}}}(\theta) 
    &\approx -\nabla \mathcal{L}(\hat{\mathbf{z}}, \theta) +N \cdot F (\theta-\theta_0) \\
    &\approx -\nabla \mathcal{L}(\hat{\mathbf{z}}, \theta),
\end{aligned}
\label{eqn:supp_first}
\end{equation}

where we assume $\theta$ is close to $\theta_0$, so the term $N \cdot F (\theta-\theta_0)$ can be omitted. Empirically, we also tried unlearning with this term added, but observed little change in performance. Then, we derive $H_{\text{unlearn}}^{\hat{\textbf{z}}}(\theta)$ as follows:

\begin{equation}
\begin{aligned}
    H_{\text{unlearn}}^{\hat{\textbf{z}}}(\theta) 
    &\approx -H_{\hat{\textbf{z}}}(\theta) +N \cdot F \\
    &\approx N \cdot F,
\end{aligned}
\label{eqn:supp_second}
\end{equation}

where $H_{\hat{\textbf{z}}}(\theta)$ is the Hessian of $ \mathcal{L}(\hat{\mathbf{z}}, \theta)$. We assume the magnitude (in Forbenius norm) of $H_{\hat{\textbf{z}}}(\theta)$ is bounded, and with a large dataset size $N$, we can approximate the Hessian $H_{\text{unlearn}}^{\hat{\textbf{z}}}(\theta)$ as $N\cdot F$ only. Incorporating Equation~\ref{eqn:supp_first} and~\ref{eqn:supp_second} into Equation~\ref{eqn:supp_newton}, we obtain our Newton update in Equation~\ref{eqn:newton_removal}:

\begin{equation}
\begin{aligned}
    \theta \leftarrow \theta + \frac{\alpha}{N} F^{-1}  \nabla\mathcal{L}(\hat{\mathbf{z}}, \theta).
\end{aligned}
\end{equation}

\subsection{Connection to influence functions}
\label{sec:supp_connect_infl}
We note that a special case of our formulation, running our update step once, with a small step size, is close to the formulation of influence functions. The difference is mainly on the linear approximation error of the loss on the training point. Starting with pretrained model $\theta_0$ and taking an infinitesimal step $\gamma$:

\begin{equation}
\begin{aligned}
    \theta_{-\hat{\mathbf{z}}} = \theta_0 + \gamma F^{-1}  \nabla\mathcal{L}(\hat{\mathbf{z}}, \theta).
\end{aligned}
\end{equation}

When we evaluate the loss of the training point $\textbf{z}$ using the unlearned model $\theta_{-\hat{\mathbf{z}}}$, we can write the loss in a linearized form around $\theta_0$, as we taking a small step:

\begin{equation}
\begin{aligned}
    \mathcal{L}(\mathbf{z}, \theta_{-\hat{\mathbf{z}}}) &\approx \mathcal{L}(\mathbf{z}, \theta_0) + \nabla\mathcal{L}(\mathbf{z}, \theta_0)^T(\theta_{-\hat{\mathbf{z}}} - \theta_0) \\
    &= \mathcal{L}(\mathbf{z}, \theta_0) + \gamma \nabla\mathcal{L}(\mathbf{z}, \theta_0)^T F^{-1}\nabla\mathcal{L}(\hat{\mathbf{z}}, \theta).
\end{aligned}
\label{eqn:supp_linear_train_loss}
\end{equation}

Now, we plug Equation~\ref{eqn:supp_linear_train_loss} into our attribution function in Equation~\ref{eqn:attribution_score}:

\begin{equation}
\begin{aligned}
    \tau(\hat{\mathbf{z}}, \mathbf{z}) &= \mathcal{L}(\mathbf{z}, \theta_{-\hat{\mathbf{z}}}) - \mathcal{L}(\mathbf{z}, \theta_0) \\
    &\approx  \gamma \nabla\mathcal{L}(\mathbf{z}, \theta_0)^T F^{-1}\nabla\mathcal{L}(\hat{\mathbf{z}}, \theta).
\end{aligned}
\end{equation}

In this special case, our method is equivalent to influence function $\nabla\mathcal{L}(\mathbf{z}, \theta_0)^T F^{-1}\nabla\mathcal{L}(\hat{\mathbf{z}}, \theta)$, after approximations. Practically, the difference between our method and influence functions is that we are taking larger, sometimes multiple steps (rather than a single, infinitesimal step), and are explicitly evaluating the loss (rather than with a linear, closed-form approximation).

\section{Implementation Details}
\label{sec:supp_impl}
\subsection{MSCOCO Models}
\myparagraph{Models trained from scratch.}
We select our source model for attribution from Georgiev~\etal~\cite{georgiev2023journey}, which is a latent diffusion model where the CLIP text encoder and VAE are exactly the ones used in Stable Diffusion v2, but with a smaller U-Net. To retrain each MSCOCO model for leave-$K$-out evaluation, we follow the same training recipe as the source model, where each model is trained with 200 epochs, a learning rate of $10^{-4}$, and a batch size of  128. \arxiv{We use the COCO 2017 training split as our training set.}

\myparagraph{Unlearning.}
To unlearn a synthesized sample in MSCOCO models, we find that running with 1 step already yields good attribution performance. We perform Newton unlearning updates with step sizes of $0.01$ and update only cross-attention KV ($W^k$, $W^v$). We find that updating cross-attention KV yields the best performance, and we later provide ablation studies on the optimal subset of layers to update. \arxiv{We sample gradients 591,435 times to estimate the diagonal Fisher information, equivalent to 5 epochs of MSCOCO training set.}

\subsection{Customized Model Benchmark}
\myparagraph{Model collection.}
As described in Section~\ref{sec:exp_abc}, we selected a subset of the dataset~\cite{wang2023evaluating} comprising of 20 models: 10 object-centric and 10 artist-style models. For all object-centric models, we select models with distinct categories. For artist-style models, we select 5 models trained from BAM-FG~\cite{ruta2021aladin} exemplars and 5 models trained from Artchive~\cite{artchive} exemplars. To speed up computation, we calculate Fisher information on Stable Diffusion v1.4, the base model of all the customized models, over the selected subset of LAION images. We then apply the same Fisher information to all customized models.

\myparagraph{Unlearning.}
We find that running 100 unlearning steps yields a much better performance than running with 1 step for this task. Moreover, updating only cross-attention KV yields a significant boost in performance in this test case. In Appendix~\ref{sec:supp_abc_additional_analysis}, we show an ablation study on these design choices. \arxiv{We sample gradients 1,000,000 times to estimate the diagonal Fisher information, where the gradients are calculated from the 100k Laion subset using Stable Diffusion v1.4.}

\subsection{Baselines.}
\myparagraph{Pixel space.} Following JourneyTRAK's implementation~\cite{georgiev2023journey}, we flatten the pixel intensities and use cosine similarity for attribution.

\myparagraph{CLIP image and text features.} We use the official ViT-B/32 model for image and text features.

\myparagraph{DINO.} We use the official ViT-B/16 model for image features.

\myparagraph{DINOv2.} We use the official ViT-L14 model with registers for image features.

\myparagraph{CLIP (AbC) and DINO (AbC).} We use the official models trained on the combination of object-centric and style-centric customized images. CLIP (AbC) and DINO (AbC) are selected because they are the best-performing choices of features.

\myparagraph{TRAK and Journey TRAK.} We adopt the official implementation of TRAK and JourneyTRAK and use a random projection dimension of 4096, the same as what they use for MSCOCO experiments.

\myparagraph{D-TRAK} \camready{We follow the best-performing hyperparameter reported in D-TRAK, using a random projection dimension of 32768 and lambda of 500. We use a single model to compute the influence score.}

\subsection{Additional Details}
\myparagraph{Horizontal flips.} \arxiv{Text-to-image models in our experiments are all trained with horizontal flips. As a result, the models are effectively also trained with the flipped version of the dataset. Therefore, we run an attribution algorithm for each training image on its original and flipped version and obtain the final score by taking the max of the two. For a fair comparison, we adopt this approach for all methods. We also find that taking the average instead of the max empirically yields similar performance.}

\myparagraph{Computational resources.}
We conduct all of our experiments on A100 GPUs. It takes around 16 hours to train an MSCOCO model from scratch, 20 hours to evaluate all training image loss, and 2 minutes to unlearn a synthesized image from a pretrained MSCOCO model. To finish all experiments on MSCOCO models, it takes around 77K GPU hours. For Customized Model Benchmark, it takes 2 hours to unlearn a synthesized image and 16 hours to track the training image loss. To finish all experiments on this benchmark, it takes around 36K GPU hours. %

\myparagraph{Licenses.}
The source model from Georgiev~\etal~\cite{georgiev2023journey} (JourneyTRAK) is released under the MIT License. The MSCOCO dataset is released under the Creative Commons Attribution 4.0 License. Stable Diffusion v2 is released under the CreativeML Open RAIL++-M License. The CLIP model is released under the MIT License. The Customized Model Benchmark is released under the Creative Commons Attribution-NonCommercial-ShareAlike 4.0 International License.

\section{Additional Analysis}

\subsection{MSCOCO Models}
\label{sec:supp_mscoco_additional_analysis}
\myparagraph{Additional plots.}
\camready{We visualize the baseline comparisons in Table~\ref{tab:big_results} as plots, where Figure~\ref{fig:coco_loss_change_baseline},\ref{fig:coco_mse_baseline},\ref{fig:coco_clip_baseline} represents loss change, MSE deviation, and CLIP deviation, respectively. We also visualize ablations studies in Table~\ref{tab:ablation_studies} as plots (Figure~\ref{fig:coco_loss_change_layer_ablation},\ref{fig:coco_mse_ablation},\ref{fig:coco_clip_ablation}).}

\camready{Densely sweeping $K$ is generally not feasible for attribution methods, as each queried synthesized image requires retraining a different set of $K$ images. However, in the plots, we report more leave-$K$-out models for random baselines, sweeping through $K =$ 500, 1000, 2000, 3000, 4000, and then 5000 to the full dataset size by increments of 2000. We use the densely swept random baselines as a reference to provide better intuition of each method's performance. For example, in Figure~\ref{fig:coco_loss_change_baseline}, removing 500 images (just $0.42\%$ of the dataset) using our method is equivalent to removing 57.5k random images (48.6$\%$ of the dataset) vs. \camready{55.8k images (47.1$\%$)} and 44.3k images (37.5$\%$) images for the best-performing baselines, \camready{D-TRAK} and DINO.}

\myparagraph{Ablation studies.}
\arxiv{We perform the following ablation studies and select hyperparameters for best performance in each test case:}

\vspace{-1mm}
\vspace{-\topsep}
\begin{itemize}[leftmargin=12pt]
  \setlength{\parskip}{0pt}
    \item \textbf{SGD (1 step):} 1 SGD step, step size $0.001$
    \item \textbf{SGD (10 steps):} 10 SGD steps, step size $0.0001$
    \item \textbf{Full weight:} 1 Newton steps, step size $0.0005$
    \item \textbf{Attention:} 1 Newton steps, step size $0.005$
    \item \textbf{Cross-attention:} 1 Newton steps, step size $0.005$
    \item \textbf{Cross-attention KV:} 1 Newton steps, step size $0.01$ (This is our final method)
\end{itemize}

The SGD step refers to the baseline of directly maximizing synthesized image loss without EWC loss regularization, as described in Section~\ref{sec:methods}.

We also compare different subsets of weights to optimize and report \camready{plots for} loss change, deviation measured by MSE, and deviation measured by CLIP similarity in Figure~\ref{fig:coco_loss_change_layer_ablation},~\ref{fig:coco_mse_ablation},~\ref{fig:coco_clip_ablation}, respectively. %

\begin{table}[t]
{\centering
{\small
\centering
\resizebox{.65\linewidth}{!}{
\begin{tabular}{lcccc}
\toprule
 & \multicolumn{2}{c}{Target Images  {\scriptsize (to forget)}}  & \multicolumn{2}{c}{Other Images  {\scriptsize (to retain)}} \\ \cmidrule(lr){2-3} \cmidrule(lr){4-5}
 & MSE ($\uparrow$) & CLIP ($\downarrow$)  & MSE ($\downarrow$) & CLIP ($\uparrow$) \\ \midrule
SGD & 0.081{\color{gray}\scriptsize±0.003} & 0.67{\color{gray}\scriptsize±0.01} & 0.033{\color{gray}\scriptsize±0.0004} & 0.83{\color{gray}\scriptsize±0.002} \\
Full weight & 0.086{\color{gray}\scriptsize±0.005} & 0.70{\color{gray}\scriptsize±0.01} & 0.039{\color{gray}\scriptsize±0.0010} & 0.86{\color{gray}\scriptsize±0.002} \\
Ours & \textbf{0.093{\color{gray}\scriptsize±0.004}} & \textbf{0.65{\color{gray}\scriptsize±0.01}} & \textbf{0.022{\color{gray}\scriptsize±0.0004}} & \textbf{0.89{\color{gray}\scriptsize±0.002}} \\ \bottomrule
\end{tabular}}}
\vspace{4pt}
\caption{\textbf{Effectiveness in Unlearning Synthesized Images.} \camready{We compare different choices of unlearning algorithms and evaluate based on whether the method can forget the target images and retain other images. We measure the performance with regenerated images' deviations via mean square error (MSE) and CLIP similarity. SGD refers to the naive baseline without EWC regularization, and full weight refers to updating on all of the weights instead of cross-attention KV.}}
\label{tab:supp_unlearn}}
\end{table}

\begin{figure}
    \centering
    \includegraphics[width=1.0\linewidth]{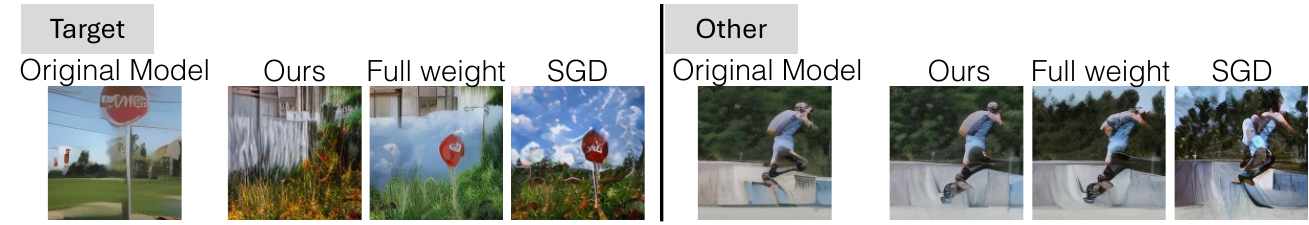}
    \caption{\textbf{Ablation for unlearning.} We find that our unlearning method (Ours) outperforms other variants (Full Weight, SGD) in terms of forgetting the target (left) while retaining other concepts (right).}
    \label{fig:supp_unlearn}
\end{figure}

\myparagraph{Effectiveness in unlearning synthesized images.}
\camready{Our attribution method relies on unlearning synthesized images, making it crucial to have an unlearning algorithm that effectively removes these images without forgetting other concepts. We analyze the performance of our unlearning algorithm itself and ablate our design choices. We construct experiments by unlearning a target synthesized image and evaluating:}

\vspace{-1mm}
\vspace{-\topsep}
\begin{itemize}[leftmargin=12pt]
  \setlength{\parskip}{0pt}
    \item \textbf{Unlearning the target image:} We measure the deviation of the regenerated image from the original model’s output—the greater the deviation, the better.
    \item \textbf{Retaining other concepts:} We generate 99 images using different text prompts and evaluate their deviations from the original model’s output—the smaller the deviation, the better.
\end{itemize}

\camready{We measure these deviations using mean square error (MSE) and CLIP similarity. We evaluate across 40 target images, with text prompts sampled from the MSCOCO validation set. We compare to the following ablations:}

\vspace{-1mm}
\vspace{-\topsep}
\begin{itemize}[leftmargin=12pt]
  \setlength{\parskip}{0pt}
    \item \textbf{SGD} refers to swapping our method’s Newton update steps (Eq.~\ref{eqn:newton_removal} ) to the naive baseline, where we run SGD steps to maximize the target loss without EWC regularization.
    \item \textbf{Full weight} refers to running Newton update steps on all of the weights instead of Cross Attn KV.
\end{itemize}

\camready{Table~\ref{tab:supp_unlearn}, along with Figure~\ref{fig:supp_unlearn}, shows the comparison. Both our regularization and weight subset optimization help unlearn the target image more effectively, without forgetting other concepts.}

\begin{table}[t]
{\small
\centering
\resizebox{1.\linewidth}{!}{
\begin{tabular}{lcccccc}
\toprule
 & \multicolumn{2}{c}{Target Images {\scriptsize (to forget)}} & \multicolumn{2}{c}{Related Images {\scriptsize (to retain)}} & \multicolumn{2}{c}{Other Images {\scriptsize (to retain)}} \\ \cmidrule(lr){2-3} \cmidrule(lr){4-5} \cmidrule(lr){6-7}
 & MSE ($\uparrow$) & CLIP ($\downarrow$) & MSE ($\downarrow$) & CLIP ($\uparrow$) & MSE ($\downarrow$) & CLIP ($\uparrow$) \\ \midrule
K=500 & 0.054{\color{gray}\scriptsize±0.004} & 0.72{\color{gray}\scriptsize±0.012} & 0.039{\color{gray}\scriptsize±0.0003} & 0.86{\color{gray}\scriptsize±0.001} & 0.041{\color{gray}\scriptsize±0.0003} & 0.79{\color{gray}\scriptsize±0.001} \\
K=1000 & 0.058{\color{gray}\scriptsize±0.004} & 0.68{\color{gray}\scriptsize±0.014} & 0.041{\color{gray}\scriptsize±0.0003} & 0.86{\color{gray}\scriptsize±0.001} & 0.041{\color{gray}\scriptsize±0.0003} & 0.79{\color{gray}\scriptsize±0.001} \\
K=4000 & 0.060{\color{gray}\scriptsize±0.004} & 0.61{\color{gray}\scriptsize±0.014} & 0.046{\color{gray}\scriptsize±0.0004} & 0.83{\color{gray}\scriptsize±0.001} & 0.041{\color{gray}\scriptsize±0.0003} & 0.79{\color{gray}\scriptsize±0.001} \\ \bottomrule
\end{tabular}}}
\vspace{1pt}
\caption{\textbf{Does leave-K-out models forget other images?} \camready{We verify that leave-$K$-model forgets concepts specific to the target query. We report deviations (MSE, CLIP) from the target image, related images that are similar to the target, and other images of unrelated concepts. We find that target images deviate more than related and other images, while other images stay almost the same. Related images's errors increase with larger $K$, but they are much smaller than target images' deviations.}}
\label{tab:supp_leave_k_out}
\end{table}

\begin{figure}
    \centering
    \includegraphics[width=0.9\linewidth]{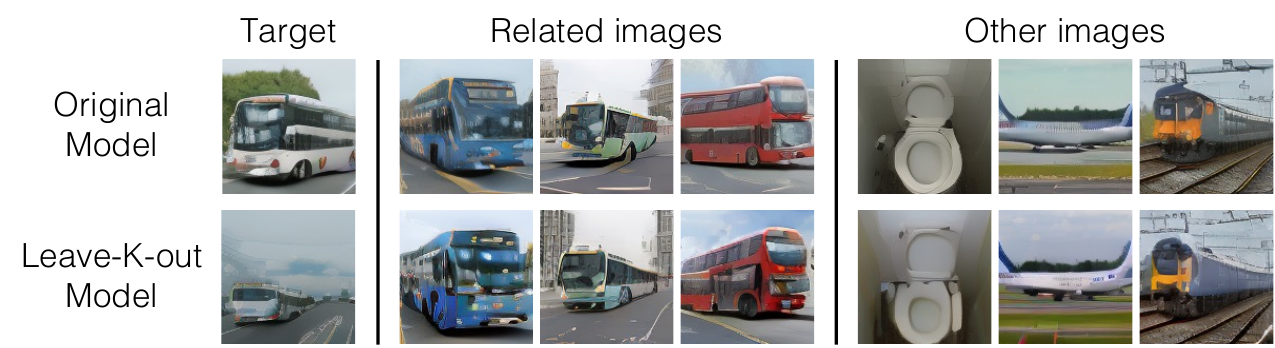}
    \caption{\textbf{Does leave-$K$-out models forget other images?} We show that leave-$K$-out model forgets the specific target (left), while retaining its generation on related images (middle) and images of other concepts (right).}
    \label{fig:bus_k500}
\end{figure}

\myparagraph{Does leave-K-out models forget other images?}
\camready{In Sec.~\ref{sec:experiment}, we show that leave-$K$-out models forget how to generate target synthesized image queries. Here we ask whether these models forget unrelated images, too. We study how much leave-$K$-out model’s generation deviates from those of the original model in three categories:}

\vspace{-1mm}
\vspace{-\topsep}
\begin{itemize}[leftmargin=12pt]
  \setlength{\parskip}{0pt}
    \item \textbf{Target images:} the attributed synthesized image. Leave-$K$-out models should forget these—the greater the deviation, the better.
    \item \textbf{Related images:} images synthesized by captions similar to the target prompt. We obtain the most similar 100 captions from the MSCOCO val set using CLIP’s text encoder. Leave-$K$-out models should not forget all of them—the smaller the deviation, the better.
    \item \textbf{Other images:} images of unrelated concepts. Prompts are 99 different captions selected from the MSCOCO val set. Leave-$K$-out models should not forget these—the smaller the deviation, the better.
\end{itemize}

\camready{In Fig~\ref{fig:bus_k500}, we find that the leave-$K$-out model “forgets” the query bus image specifically while retaining other buses and other concepts. In Table~\ref{tab:supp_leave_k_out}, we quantitatively report deviations using mean square error (MSE) and CLIP similarity, where we evaluate 40 pairs of target images and leave-K-out models. We observe that target images have larger MSE and lower CLIP similarity than related images and other images. Also, as the number of removed influential images ($K$) increases, the target image error increases rapidly while other images stay almost the same. Interestingly, related images’ errors increase with larger $K$, but the errors are still much smaller than those of target images. As $K$ increases, the group of influential images can start affecting other related concepts.}

\begin{table}[]
{\small
\centering
\resizebox{1.\linewidth}{!}{
\begin{tabular}{lcccccccccc}
\toprule
 &  & \multicolumn{3}{c}{$\Delta \mathcal{L}(\hat{\mathbf{z}}, \theta)$ $\uparrow$ {\footnotesize(x$10^{-3}$)}} & \multicolumn{3}{c}{$\Delta G_\theta(\epsilon, \mathbf{c})$ (MSE) $\uparrow$ {\footnotesize(x$10^{-2}$)}} & \multicolumn{3}{c}{$\Delta G_\theta(\epsilon, \mathbf{c})$ (CLIP) $\downarrow$ {\footnotesize(x$10^{-1}$)}} \\ \cmidrule(lr){3-5}\cmidrule(lr){6-8}\cmidrule(lr){9-11}
\multirow{-2}{*}{Method} & \multirow{-2}{*}{Loss} & \multicolumn{1}{c}{K=500} & \multicolumn{1}{c}{K=1000} & \multicolumn{1}{c}{K=4000} & \multicolumn{1}{c}{K=500} & \multicolumn{1}{c}{K=1000} & \multicolumn{1}{c}{K=4000} & \multicolumn{1}{c}{K=500} & \multicolumn{1}{c}{K=1000} & \multicolumn{1}{c}{K=4000} \\ \midrule
TRAK & $||\epsilon_\theta(\mathbf{x}_t, t) - \epsilon||^2$ & 5.2{\color{gray}\scriptsize±0.14} & 5.8{\color{gray}\scriptsize±0.16} & 7.1{\color{gray}\scriptsize±0.16} & 4.7{\color{gray}\scriptsize±0.21} & 4.7{\color{gray}\scriptsize±0.24} & 4.7{\color{gray}\scriptsize±0.20} & 7.6{\color{gray}\scriptsize±0.09} & 7.6{\color{gray}\scriptsize±0.09} & 7.5{\color{gray}\scriptsize±0.09} \\
D-TRAK & $||\epsilon_\theta(\mathbf{x}_t, t)||^2$ & 5.4{\color{gray}\scriptsize±0.16} & 6.6{\color{gray}\scriptsize±0.22} & 9.6{\color{gray}\scriptsize±0.33} & 5.9{\color{gray}\scriptsize±0.24} & 6.4{\color{gray}\scriptsize±0.25} & 7.8{\color{gray}\scriptsize±0.30} & 7.3{\color{gray}\scriptsize±0.10} & 7.1{\color{gray}\scriptsize±0.09} & 6.4{\color{gray}\scriptsize±0.09} \\ \hdashline
Ours & $||\epsilon_\theta(\mathbf{x}_t, t) - \epsilon||^2$ & 5.6{\color{gray}\scriptsize±0.16} & 6.7{\color{gray}\scriptsize±0.22} & 9.8{\color{gray}\scriptsize±0.32} & 5.1{\color{gray}\scriptsize±0.21} & 5.7{\color{gray}\scriptsize±0.24} & 6.1{\color{gray}\scriptsize±0.22} & 7.3{\color{gray}\scriptsize±0.09} & 7.0{\color{gray}\scriptsize±0.09} & 6.4{\color{gray}\scriptsize±0.10} \\
Ours (D-TRAK) & $||\epsilon_\theta(\mathbf{x}_t, t)||^2$ & 4.4{\color{gray}\scriptsize±0.14} & 5.1{\color{gray}\scriptsize±0.20} & 7.6{\color{gray}\scriptsize±0.36} & 4.7{\color{gray}\scriptsize±0.23} & 4.9{\color{gray}\scriptsize±0.24} & 6.0{\color{gray}\scriptsize±0.28} & 7.6{\color{gray}\scriptsize±0.09} & 7.4{\color{gray}\scriptsize±0.09} & 6.3{\color{gray}\scriptsize±0.09} \\ \bottomrule
\end{tabular}}}
\vspace{1pt}
\caption{\textbf{Choice of loss function.} \camready{We study different choices of loss. While D-TRAK~\cite{zheng2023intriguing} empirically showed that changing the diffusion loss to a square loss increases performance drastically for influence function, we find that this loss change does not grant performance gain for our formulation. We report $\Delta \mathcal{L}(\hat{\mathbf{z}}, \theta)$ and $\Delta G_\theta(\epsilon, \mathbf{c})$ as in Table~\ref{tab:big_results}.}}
\label{tab:supp_dtrak_analysis}
\end{table} 
\myparagraph{Applying the same loss function changes as D-TRAK.}
\camready{As mentioned in Sec.~\ref{sec:exp_leave_k_out}, D-TRAK yields a significant performance gain by changing the loss function in TRAK from a denoising loss ($\mathbb{E}[\| \epsilon_{\theta}(\mathbf{x}_t, \mathbf{c}, t) - \epsilon\|^2]$) to a square loss ($\mathbb{E}[\| \epsilon_{\theta}(\mathbf{x}_t, \mathbf{c}, t)\|^2]$). In Table~\ref{tab:supp_dtrak_analysis}, we find that applying the same loss change to our method leads to worse performance. 
}

\myparagraph{Additional results.}
We provide more attribution results in Figure~\ref{fig:supp_mscoco_baseline} and more results on leave-$K$-out models in Figure~\ref{fig:supp_mscoco_leave_k_out}.

\begin{figure}
    \centering
    \begin{tabular}{cc}
        \includegraphics[width=0.5\linewidth]{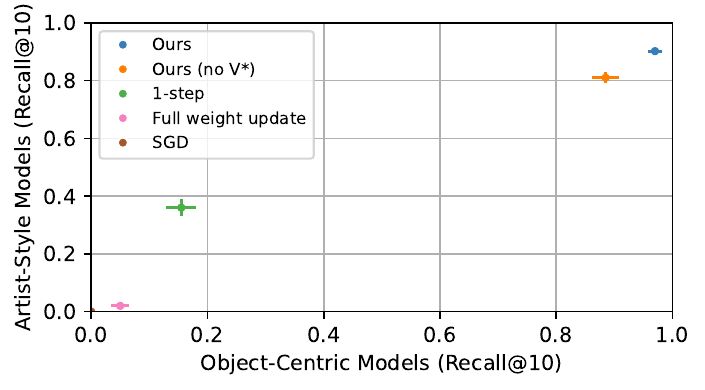} & 
        \includegraphics[width=0.5\linewidth]{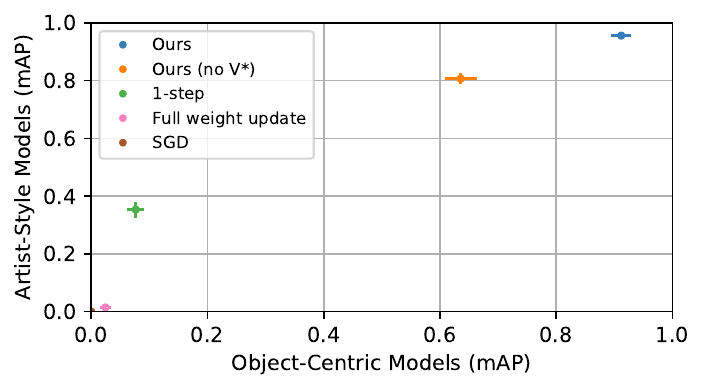} 
    \end{tabular}
    \vspace{-15pt}
    \caption{\textbf{Ablation studies for Customized Model Benchmark.} We report evaluation on the Customized Model Benchmark in the same fashion as in Figure~\ref{fig:abc_baseline}. We find that training with multiple steps, updating a selected subset of weights, and regularizing unlearning via Fisher information is crucial to this task. Additionally, we test a version where we apply our algorithm without the special token {\menlo V$^{*}$}. While it reduces performance, it still performs well in overall.}
    \label{fig:abc_ablation}
\end{figure}

\subsection{Customized Model Benchmark}
\myparagraph{Ablation studies.}
\label{sec:supp_abc_additional_analysis}
We perform the following ablation studies and select hyperparameters for best performance in each test case:

\vspace{-1mm}
\vspace{-\topsep}
\begin{itemize}[leftmargin=12pt]
  \setlength{\parskip}{0pt}
    \item \textbf{Cross-attention KV (100 steps):} 100 Newton steps, step size 0.1 (denoted as \textbf{Ours} in Figure~\ref{fig:abc_ablation})
    \item \textbf{Cross-attention KV (1 step):} 1 Newton step, step size 10
    \item \textbf{Full weight (100 steps):} 100 Newton steps, step size $5 \times 10^{-5}$
    \item \textbf{Cross-attention KV, SGD step (100 steps):} 100 SGD step, step size $0.01$
\end{itemize}

Again, the SGD step refers to the baseline of directly maximizing synthesized image loss without EWC loss regularization, as described in Section~\ref{sec:methods}.

We report the result of our ablation studies in Figure~\ref{fig:abc_ablation}. Our findings indicate that for this test case, selecting a small subset of weights (i.e., cross-attention KV) combined with multiple unlearning steps (100 steps) is crucial for effective attribution. We hypothesize that stronger regularization is necessary for unlearning in larger-scale models, and that such models benefit more from numerous smaller unlearning steps rather than fewer, larger steps to achieve a better optimization.

Customized models in this benchmark associate the exemplar with a special token {\menlo V$^{*}$}, which is also used for generating synthesized images. Our method involves forgetting the synthesized image associated with its text prompt, so by default, we tested it with {\menlo V$^{*}$} included. Meanwhile, we also evaluated our method without {\menlo V$^{*}$} in the prompts. Figure~\ref{fig:abc_baseline} shows that removing {\menlo V$^{*}$} reduces performance, but the method still performs well overall.

\section{Efficiency Comparison with Other Methods}
\label{sec:supp_runtime}
\camready{We compare our method’s efficiency with other methods. Below, we briefly overview each method:}

\vspace{-1mm}
\vspace{-\topsep}
\begin{itemize}[leftmargin=12pt]
  \setlength{\parskip}{0pt}
    \item \textbf{Our unlearning method:} we obtain a model that unlearns the synthesized image query and checks the loss increases for each training image after unlearning.
    \item \textbf{TRAK, JourneyTRAK:} precompute randomly-projected loss gradients for each training image. Given a synthesized image query, the authors first obtain its random-projected loss gradient and then match it with the training gradients using the influence function. For good performance, both methods require running influence function on multiple pre-trained models (e.g., 20 for MSCOCO).
    \item \textbf{D-TRAK:} \camready{The runtime complexity is similar to that of TRAK, but it is running the influence function on a single pre-trained model.}
    \item \textbf{Feature similarity:} standard image retrieval pipeline. Precompute features from the training set. Given a synthesized image query, we compute feature similarity for the entire database.
\end{itemize}

\camready{We compare the efficiency of each method.}

\myparagraph{Feature similarity is the most run-time efficient} \camready{since obtaining features is faster than obtaining losses or gradients from a generative model. However, feature similarity does not leverage knowledge of the generative model. Our method outperforms various feature similarity methods (Table~\ref{tab:big_results}).}

\myparagraph{Our method is more efficient than TRAK/JourneyTRAK/D-TRAK in precomputation.} \camready{Our method’s precomputation is much more efficient runtime and storage-wise compared to TRAK, JourneyTRAK, and D-TRAK. Our method only requires computing and storing loss values of training images from a single model. On the other hand, TRAK-based methods require pre-training extra models (e.g., 20) from scratch, and precomputing/storing loss gradients of training images from those models.}

\myparagraph{Our method is less efficient when estimating attribution score.} \camready{TRAK-based methods obtain attribution scores by taking a dot product between synthesized image gradient and the stored training gradient features. The methods require calculating and averaging such dot product scores from the 20 pretrained models. On the other hand, as acknowledged in our limitations (Sec.~\ref{sec:discussion}), though our model unlearns efficiently (e.g., only 1-step update for MSCOCO), getting our attribution score involves estimating the loss on the training set, which is less efficient than dot product search. Tradeoff-wise, our method has a low storage requirement at the cost of higher runtime.}

\camready{Our main objective is to push the envelope on the difficult challenges of attribution performance. Improving computation efficiency of attribution is a challenge shared across the community~\cite{park2023trak,grosse2023studying}, and we leave it to future work.}

\begin{figure}
    \centering
    \includegraphics[width=\linewidth]{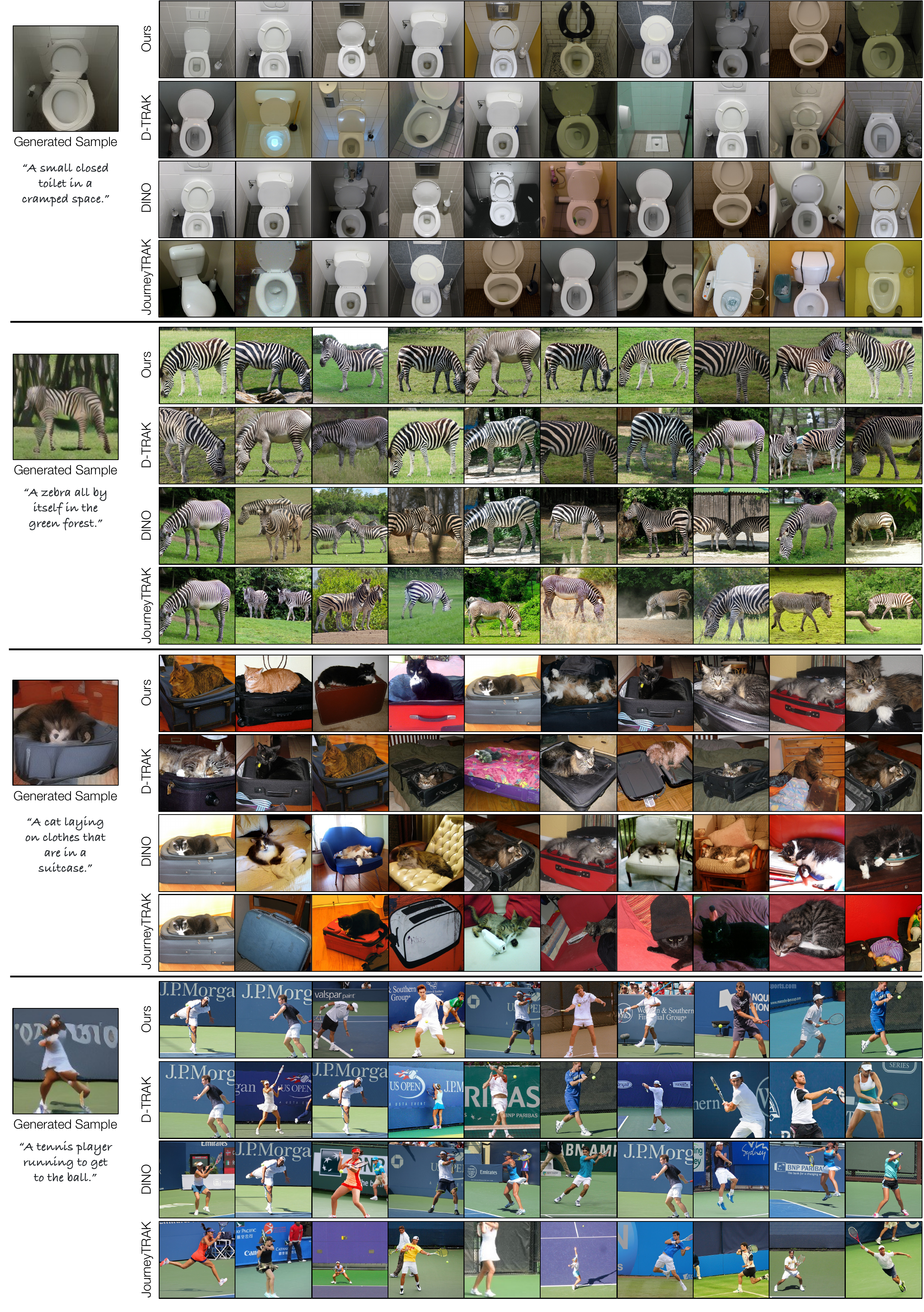}
    \caption{\textbf{Additional attribution results for MSCOCO models.} This is an extension of Figure~\ref{fig:qual_coco_baseline} in the main paper. Our method identifies images with similar poses and visual attributions to the query image. Importantly, in Figure~\ref{fig:supp_mscoco_leave_k_out}, we verify that leaving these images out of training corrupts the ability of the model to synthesize the queried images.}
    \label{fig:supp_mscoco_baseline}
\end{figure}

\begin{figure}
    \centering
    \includegraphics[width=\linewidth]{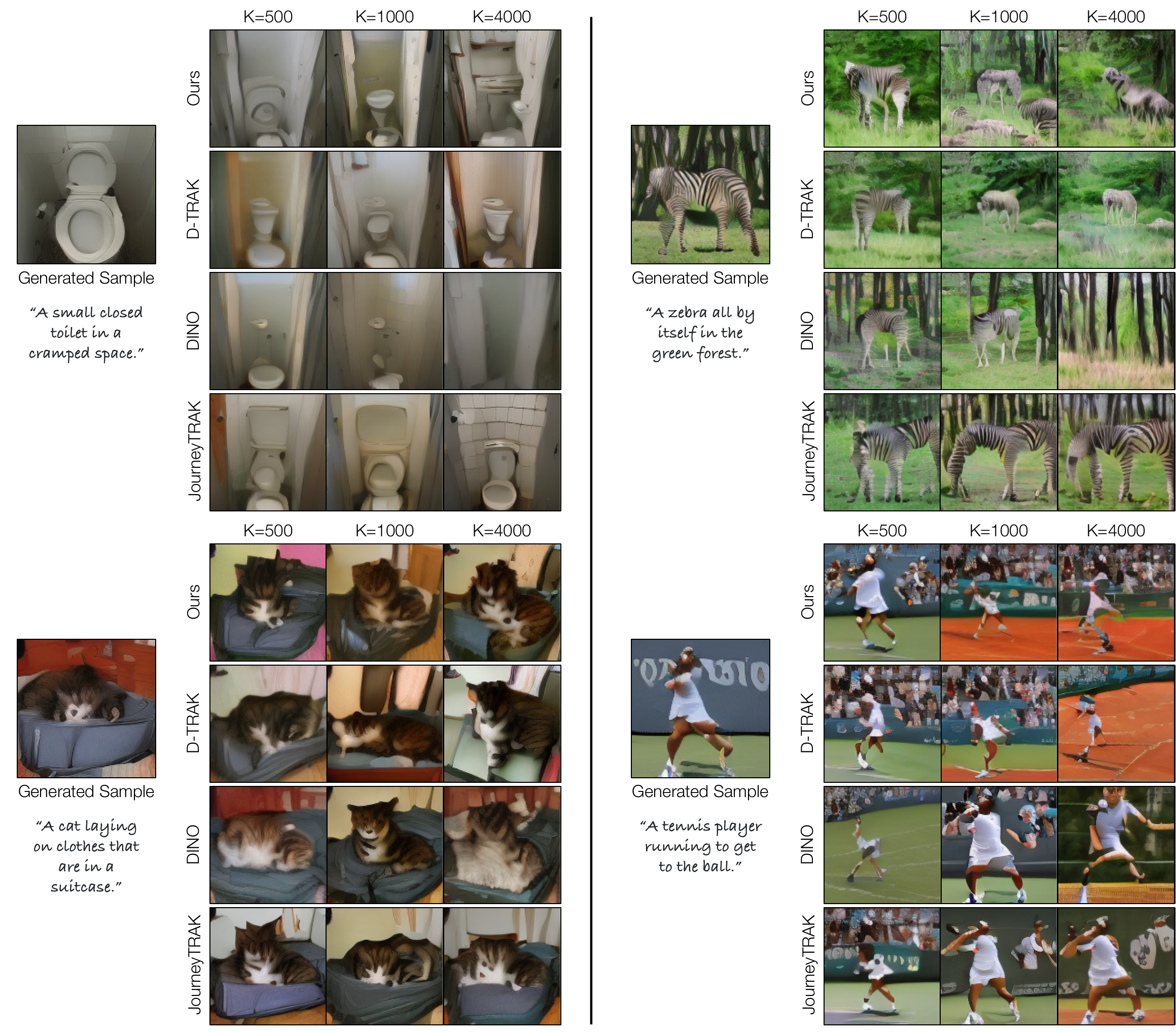}
    \caption{\textbf{Additional leave-$K$-out model results for MSCOCO models.} This is an extension of Figure~\ref{fig:qual_coco_leave_k} in the main paper, showing the results from removing top-$K$ influential images from different algorithms, retraining, and attempting to regenerate a synthesized sample. The influential images for these examples are shown in Figure~\ref{fig:supp_mscoco_baseline}. Our method consistently destroys the synthesized examples, verifying that our method is identifying the critical influential images.}
    \label{fig:supp_mscoco_leave_k_out}
\end{figure}

\begin{figure}[h]
    \centering
    \begin{tabular}{c}
         \includegraphics[width=\linewidth]{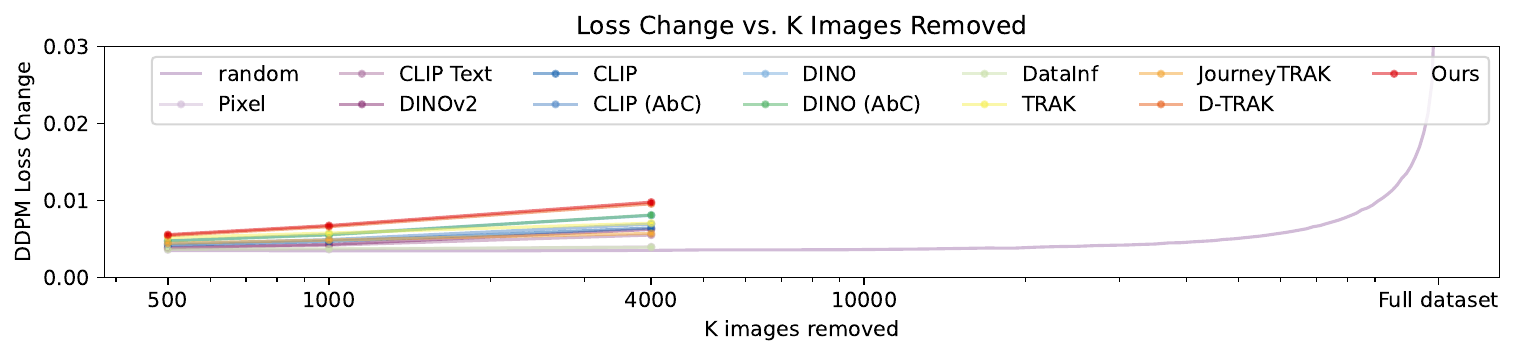}
          \vspace{-2mm}
         \\
         \includegraphics[width=\linewidth]{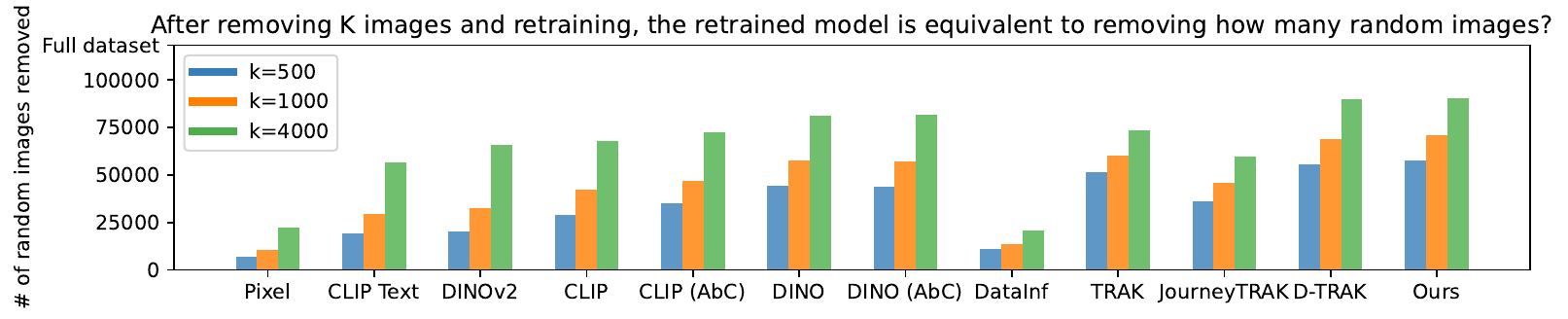}
    \end{tabular}
    \caption{\textbf{Evaluating loss changes in leave-$K$-out models.} Given a synthesized image  $\hat{\mathbf{z}}$, we train leave-$K$-out models for each of the attribution methods and track $\Delta \mathcal{L}(\hat{\mathbf{z}}, \theta)$, the increase in loss change.
    \textbf{(Top)} We plot DDPM loss change versus $K$. The loss change becomes more significant with a larger $K$ value, and our method (brown) yields the largest loss change for every $K$, indicating our method more successfully identifies the influential images. The purple curve represents models where random images are removed, serving as a reference. We show error bars representing 1 standard error in this plot (which are small and negligible). \textbf{(Bottom)} For each method, we show the equivalent number of random images needed to achieve a loss change.}
    \label{fig:coco_loss_change_baseline}
\end{figure}

\begin{figure}
    \centering
    \begin{tabular}{c}
         \includegraphics[width=\linewidth]{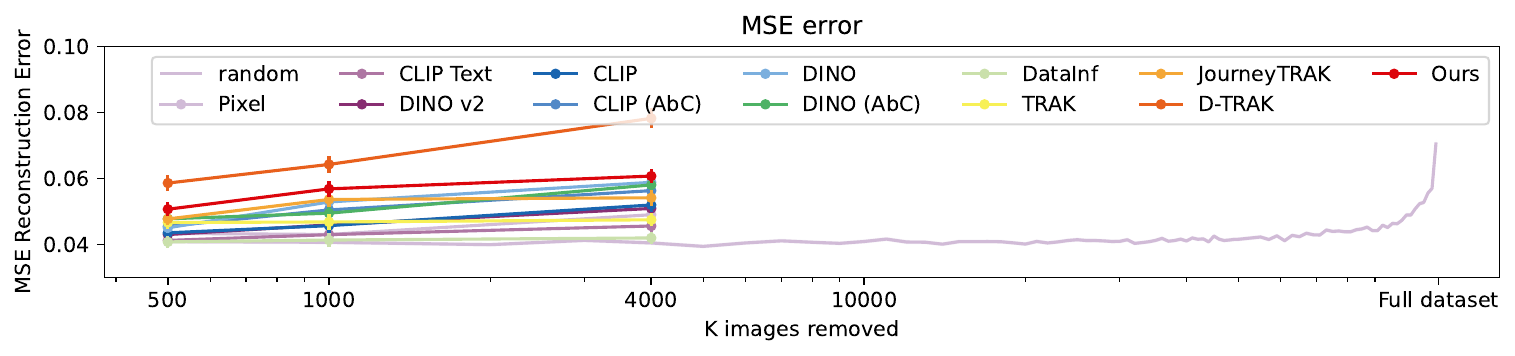} \\
         \includegraphics[width=\linewidth]{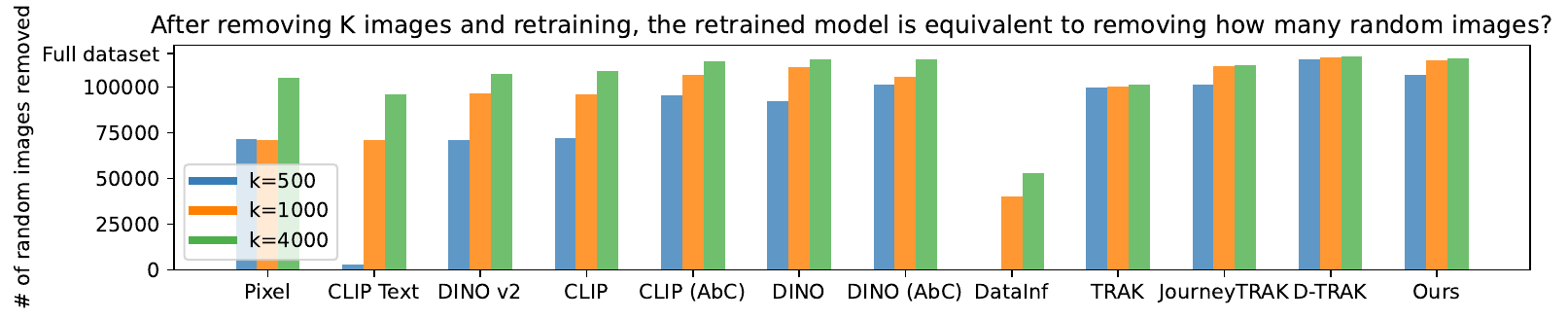}
    \end{tabular}
    \caption{\textbf{Deviation of generated output in leave-$K$-out models, in mean square error (MSE).} We report the deviation of generated output in the same fashion as in Figure~\ref{fig:coco_loss_change_baseline}. The higher the MSE, the better since more deviation indicates that the $K$ images removed are more influential.}
    \label{fig:coco_mse_baseline}
\end{figure}

\begin{figure}
    \centering
    \begin{tabular}{c}
         \includegraphics[width=\linewidth]{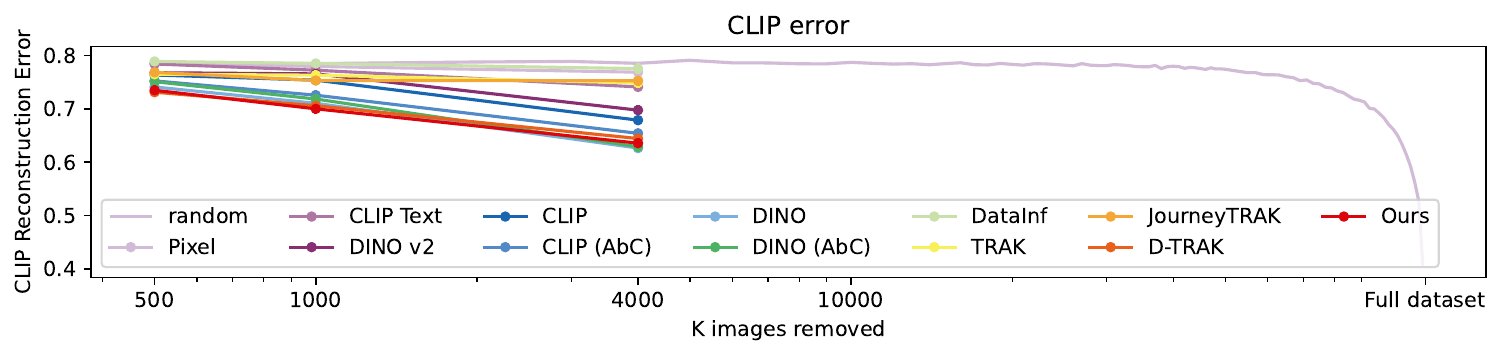} \\
         \includegraphics[width=\linewidth]{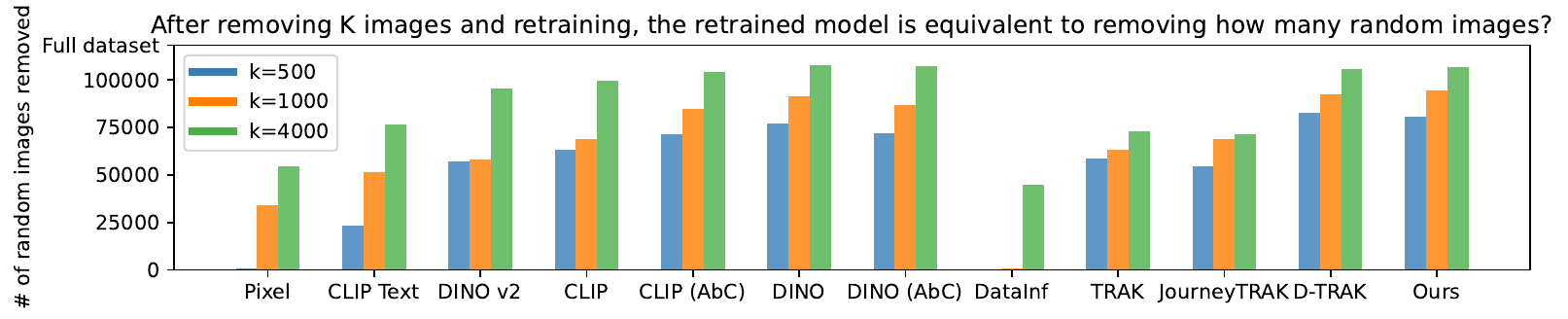}
    \end{tabular}
    \caption{\textbf{Deviation of generated output in leave-$K$-out models, in CLIP similarity.} We report the deviation of generated output in the same fashion as in Figure~\ref{fig:coco_loss_change_baseline}. A lower CLIP similarity in leave-$K$-out models indicates a better attribution algorithm.}
    \label{fig:coco_clip_baseline}
\end{figure}

\begin{figure}
    \centering
    \begin{tabular}{c}
         \includegraphics[width=\linewidth]{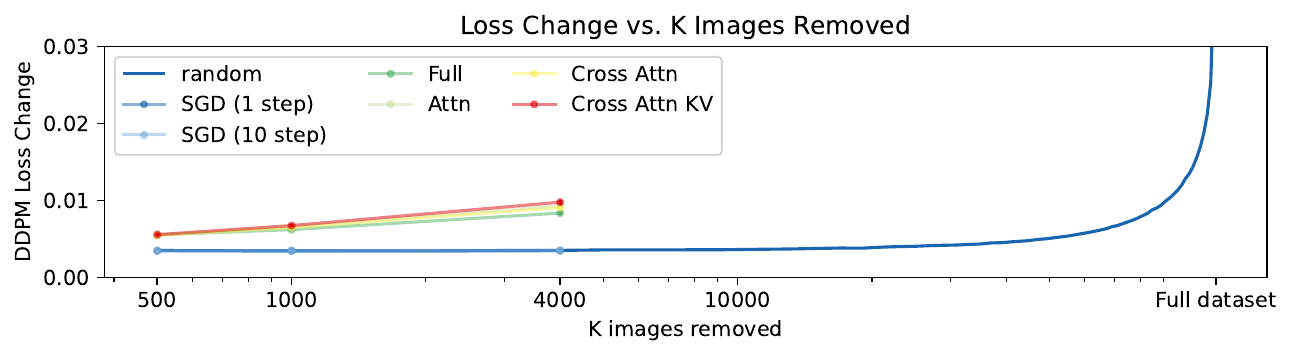} \\
         \includegraphics[width=\linewidth]{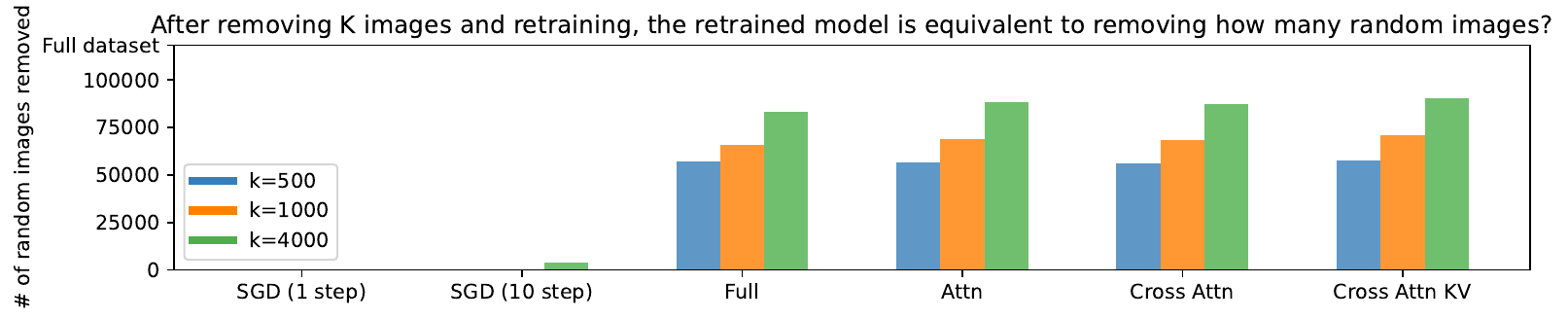}
    \end{tabular}
    \caption{\textbf{Ablation studies for attributing MSCOCO models.} We report the change in synthesized image loss in leave-$K$-out models in the same fashion as in Figure~\ref{fig:coco_loss_change_baseline} in the main paper. \arxiv{We compare four different sets of weights to unlearn (full, attention layers, cross-attention layers, and cross-attention $W^k$, $W^v$), and we find that cross-attention $W^k$, $W^v$ outperforms other configurations. We also evaluate a naive variation where we apply SGD to maximize the synthesized image's loss, and we find that updating with 1 steps or 10 steps both perform only at chance (random baseline).}}
    \label{fig:coco_loss_change_layer_ablation}
\end{figure}

\begin{figure}
    \centering
    \begin{tabular}{c}
         \includegraphics[width=\linewidth]{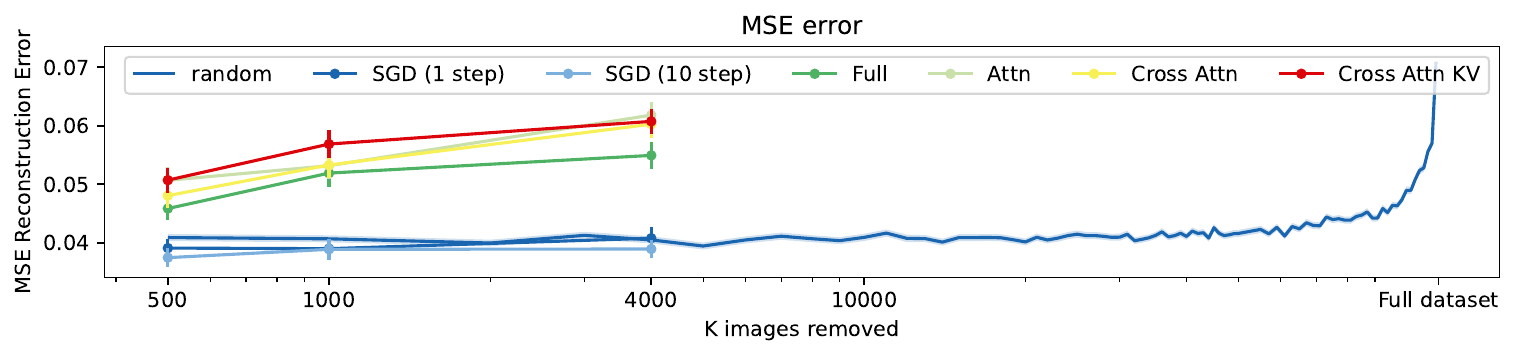} \\
         \includegraphics[width=\linewidth]{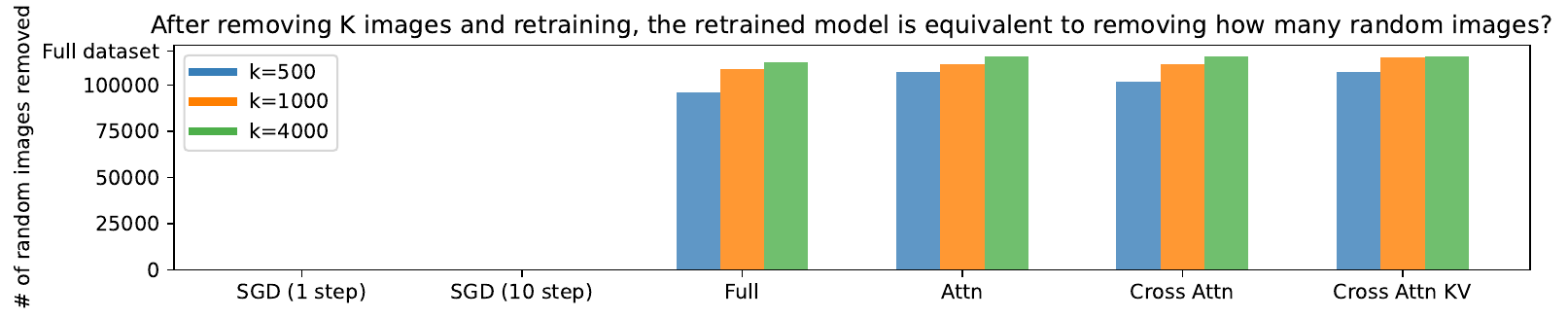}
    \end{tabular}
    \vspace{-15pt}
    \caption{\textbf{Ablation studies for attributing MSCOCO models.} We report the deviation of generated output in leave-$K$-out models in mean square error (MSE) in the same fashion as in Figure~\ref{fig:coco_mse_baseline}. We find that the trend of each ablation follows that of Figure~\ref{fig:coco_loss_change_layer_ablation}.}.
    \label{fig:coco_mse_ablation}
\end{figure}

\begin{figure}
    \centering
    \begin{tabular}{c}
         \includegraphics[width=\linewidth]{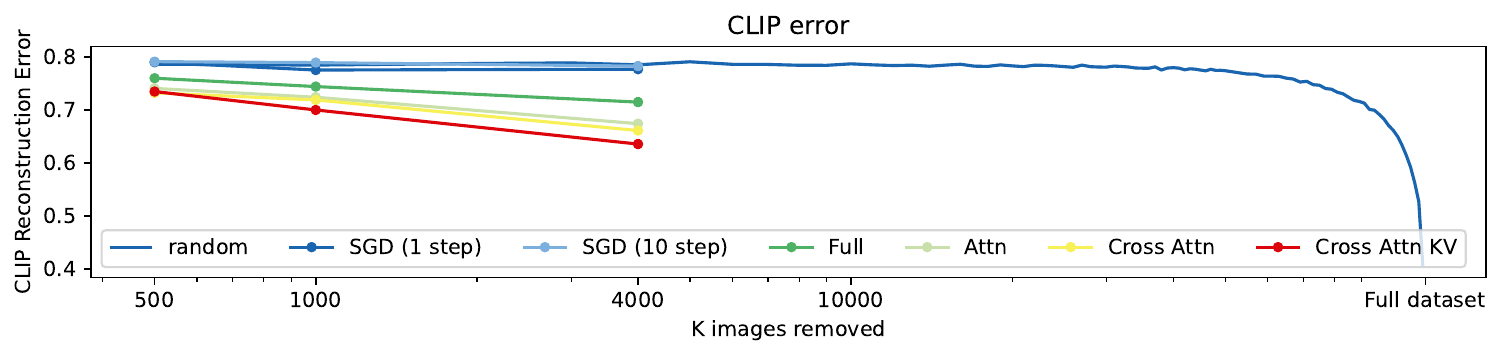} \\
         \includegraphics[width=\linewidth]{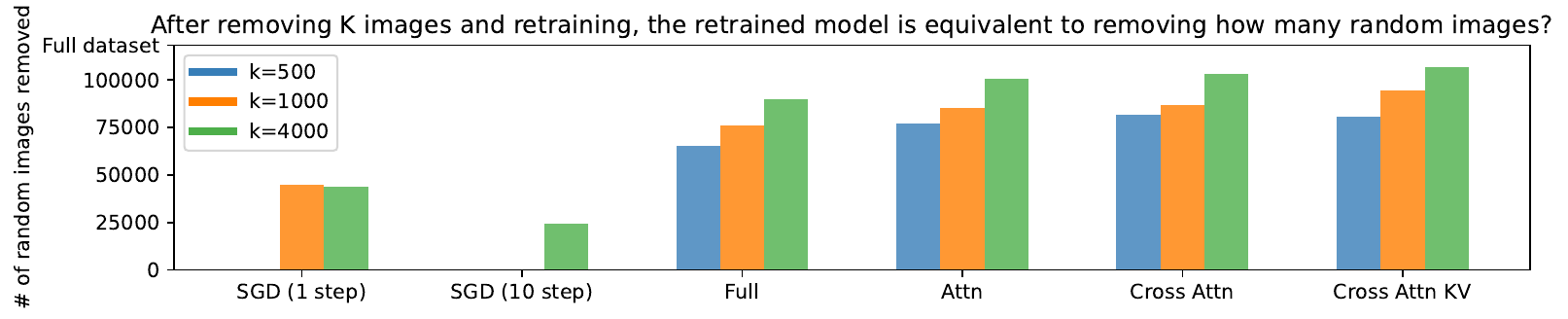}
    \end{tabular}
    \vspace{-15pt}
    \caption{\textbf{Ablation studies for attributing MSCOCO models.} We report the deviation of generated output in leave-$K$-out models in CLIP similarity in the same fashion as in Figure~\ref{fig:coco_clip_baseline}. We find that the trend of each ablation follows that of Figure~\ref{fig:coco_loss_change_layer_ablation}.}
    \label{fig:coco_clip_ablation}
\end{figure}

\vspace{-10pt}
\section{Change log.}
\vspace{-5pt}
\myparagraph{v1} Initial release.

\myparagraph{v2} NeurIPS 2024 camera ready. Figure~\ref{fig:teaser} is updated with more training images. Section~\ref{sec:experiment}, Figures~\ref{fig:qual_coco_baseline},~\ref{fig:qual_coco_leave_k},~\ref{fig:qual_abc_baseline},~\ref{fig:abc_baseline},~\ref{fig:supp_mscoco_baseline},~\ref{fig:supp_mscoco_leave_k_out},~\ref{fig:coco_loss_change_baseline},~\ref{fig:coco_mse_baseline}, and~\ref{fig:coco_clip_baseline} are edited to incorporate the two additional baselines. Tables~\ref{tab:big_results} and~\ref{tab:ablation_studies} are added to clarify the experimental results. Additional analyses in the unlearning and leave-K-out models are added in Appendix~\ref{sec:supp_mscoco_additional_analysis}, Figures~\ref{fig:supp_unlearn},~\ref{fig:bus_k500}, and Tables~\ref{tab:supp_unlearn},~\ref{tab:supp_leave_k_out}. Additional analysis of D-TRAK is added in Appendix~\ref{sec:supp_mscoco_additional_analysis}, and efficiency comparison is added in Appendix~\ref{sec:supp_runtime}. Errors corrected in Appendix~\ref{sec:supp_connect_infl}.

\myparagraph{v3} Update acknowledgments.

 \clearpage %

\end{document}